\documentclass[11pt,preprint,onecolumn,nocopyrightspace]{sigplanconf}
\usepackage[normalem]{ulem}
\usepackage{amsfonts}
\usepackage{amsmath}
\usepackage{amssymb}
\usepackage{amsthm}
\usepackage{tabularx}

\usepackage[utf8]{inputenc}
\usepackage{enumerate}
 \usepackage{array}
 \usepackage{url}
\usepackage{tikz}
\usepackage{xcolor}
\usepackage{balance}
\usepackage{listings}
\newtheorem{definition}{Definition}

\newtheorem{theorem}{Theorem}

\newcolumntype{C}{ >{\centering\arraybackslash} m{.25\textwidth} }
\newcolumntype{D}{ >{\centering\arraybackslash} m{.7\textwidth} }

\newcommand{\mypar}[1]{\vspace{0.1cm} \noindent {\em \textbf{#1}\/}}

\usepackage{xcolor}

\DeclareMathOperator*{\argmin}{arg\,min}

\begin{document}

\setlength{\pdfpageheight}{\paperheight}
\setlength{\pdfpagewidth}{\paperwidth}

\title{TC-DTW: Accelerating Multivariate Dynamic Time Warping Through Triangle Inequality and Point Clustering}

\authorinfo{Daniel S. Shen${}^*$ \and Min Chi${}^+$}
{${}^*$William G. Enloe High School, Raleigh, NC, USA\\
${}^+$North Carolina State University, Raleigh, NC, USA}
{danielsongshen@gmail.com, mchi@ncsu.edu}

\maketitle

\begin{abstract}
Dynamic time warping (DTW) plays an important role in analytics on time series. Despite the large body of research on speeding up {\em univariate} DTW, the method for {\em multivariate} DTW has not been improved much in the last two decades. The most popular algorithm used today is still the one developed seventeen years ago.  This paper presents a solution that, as far as we know, for the first time consistently outperforms the classic multivariate DTW algorithm across dataset sizes, series lengths, data dimensions, temporal window sizes, and machines. The new solution, named TC-DTW, introduces Triangle Inequality and Point Clustering into the algorithm design on lower bound calculations for multivariate DTW. In experiments on DTW-based nearest neighbor finding, the new solution avoids as much as 98\% (60\% average) DTW distance calculations and yields as much as 25$\times$ (7.5$\times$ average) speedups. 

\end{abstract}






\section{Introduction}

\begin{figure}
    \centering
    \begin{tabular}{cc}
        \begin{minipage}{0.4\textwidth}
        \centering
        \includegraphics[width=.7\textwidth]{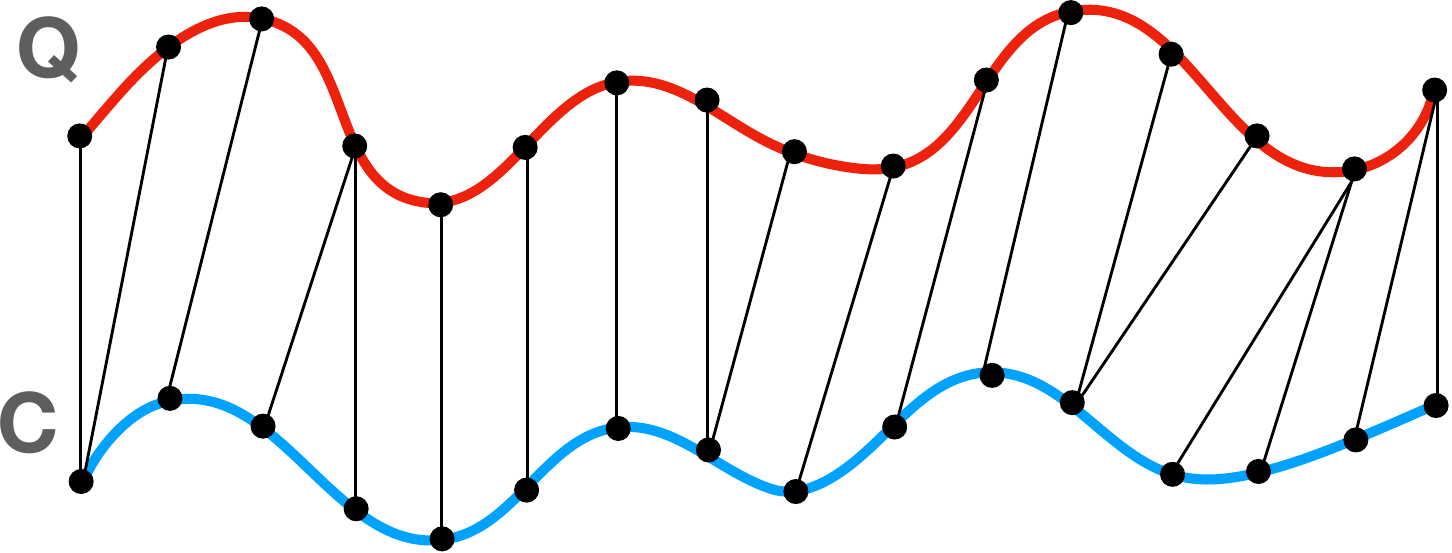}\\
        (a) DTW is to find the best mapping\\[.3cm]
        \includegraphics[width=.5\textwidth]{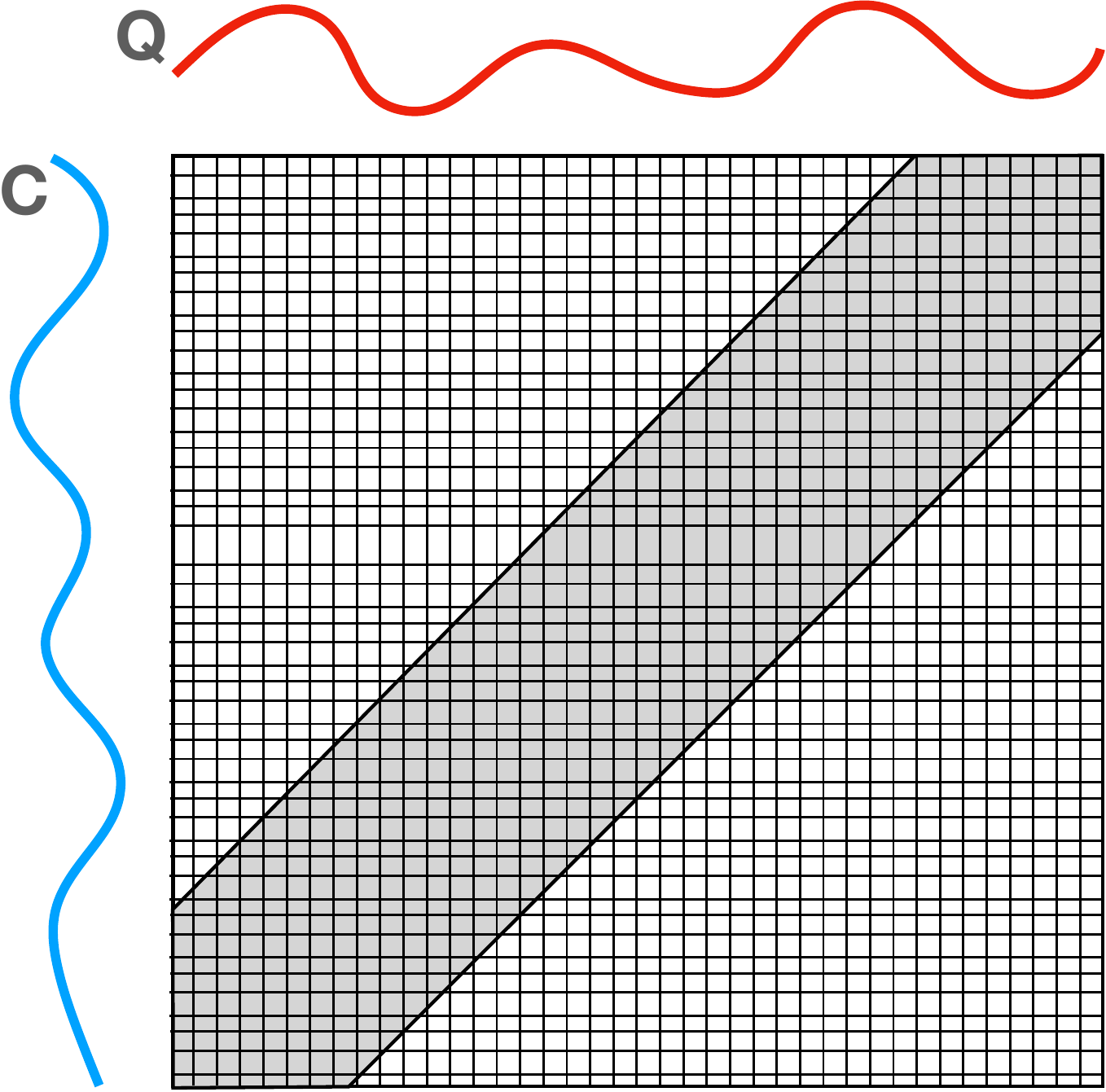}\\
        (b) Difference matrix\\[.3cm]
        \includegraphics[width=.9\textwidth]{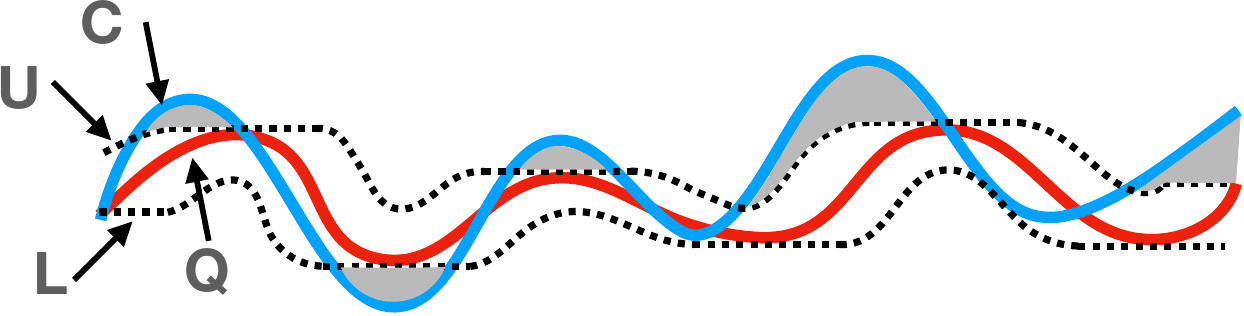}\\
        (c) Bounding envelope of a 1D series and the LB\_Keogh lower bounds
        \end{minipage}     &
    \begin{minipage}{0.55\textwidth}
            \flushright\includegraphics[width=0.35\textwidth]{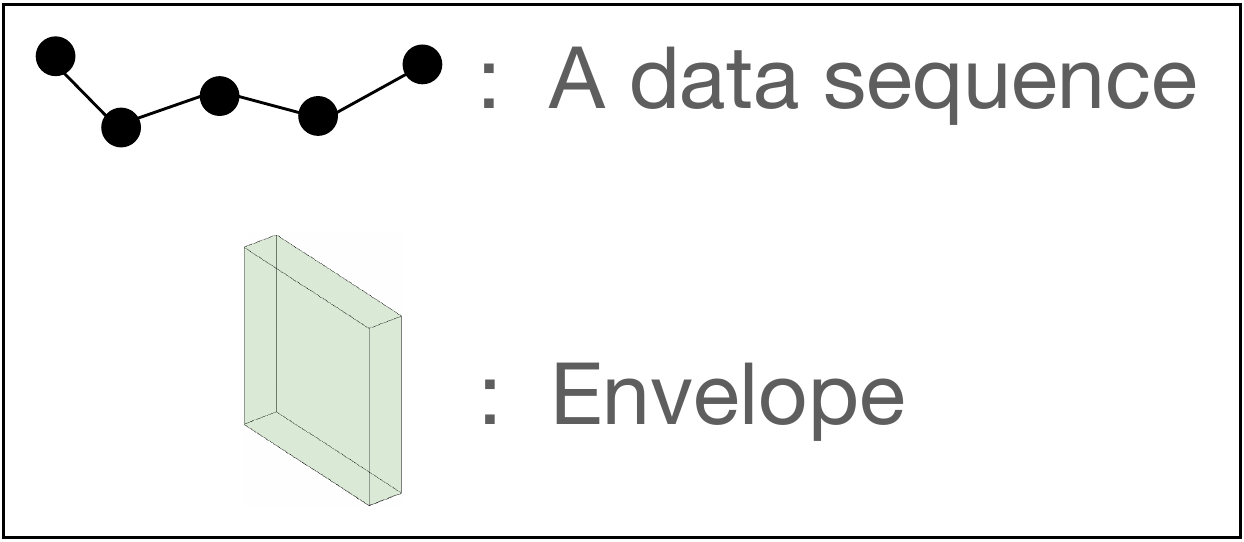}\\
        \includegraphics[width=1\textwidth]{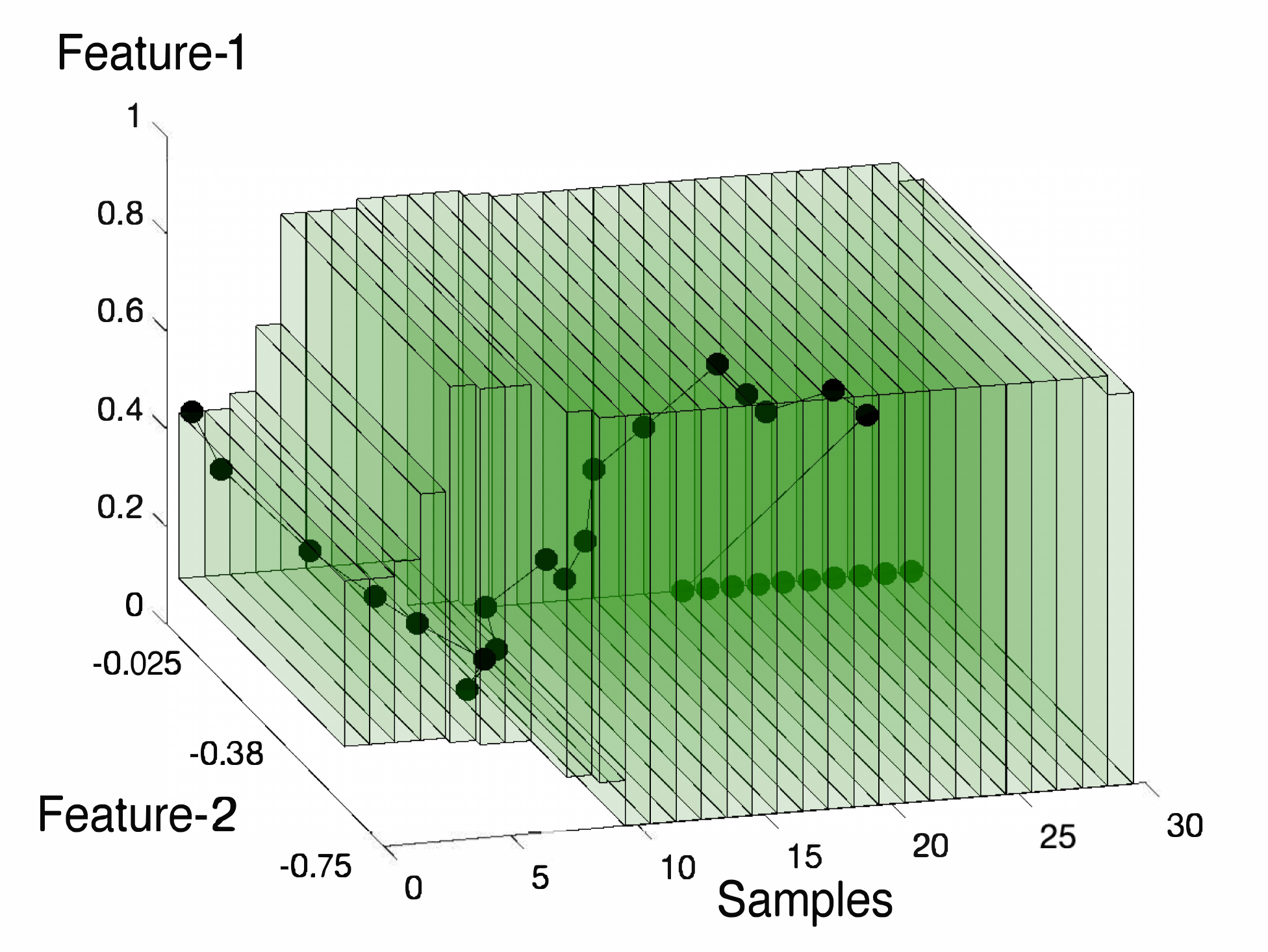}\\
        (d) Bounding envelope of a 2D series
    \end{minipage}
    \end{tabular}
    \caption{(a) DTW is about finding the mapping between the points on two sequences that minimizes the accumulated differences between the corresponding points. (b) The difference matrix between two sequences Q and C, and the Sakoe-Chuba Band. (c) The bounding envelope of sequence Q is used for computing the Keogh lower bounds (summation of the shaded area) of the DTW distance between Q and C. U and L: two series that form an envelope of Q. (d) The bounding envelope on a 2D series is even looser due to the stretching effects on both dimensions.}
    \label{fig:windowDTW}
\end{figure}

Temporal data analytics is important for many domains that have temporal data. One of its fundamental questions is calculating the similarity between two time series, due to temporal distortions. In the medical records of two patients, for instance, the times when their checks were done often differ; the points on two motion tracking series often differ in the arrival times of their samples. Consequently, it is important to find the proper alignment of the points in two time series before their similarity (or differences) can be computed. In such an alignment, as illustrated in Figure~\ref{fig:windowDTW}(a), one point may match with one or multiple points in the other series.

The most influential method to find the proper alignments is Dynamic Time Warping (DTW). It uses dynamic programming to find the best mapping, that is, an alignment that makes the accumulated differences of the aligned points in the two series minimum. Such an accumulated difference is called {\em DTW distance}. Experimental comparisons of DTW to most other distance measures on many datasets have concluded that DTW almost always outperforms other measures~\cite{Wang+:DMKD2013}. It hence has been used as one of the most preferable measures in pattern matching tasks for time series data~\cite{Chadwick+:DEST2011,Adams+:ISMIR05,Alon+:PAMI2009,Laerhoven+:ICMLA2009,Dupasquier+:Chaos2011,Huber+:MVA2011,Inglada+:sense2015,Rak+:TKDD2013}.


DTW remains popular even in the era of deep learning, for several reasons. DTW-based temporal data analytics are still widely used for its advantages in better interpretability over deep neural networks. Moreover, there is also interest in combining DTW with deep learning by, for instance, using DTW as part of the loss function. Studies have shown that DTW, compared to conventional Euclidean distance loss, leads to a better performance in a number of applications~\cite{Cuturl+:arxiv2017} through softmin~\cite{Chang+:arxiv2019,Shah+:CODS16}.  A more recent way to combine DTW and deep learning is to use DTW as a replacement of kernels in convolutional neural networks, creating DTWNet that shows improved inference accuracy~\cite{Cai+:NIPS2019}.

Even with dynamic programming, DTW is still time-consuming to compute. On its own, DTW has an $O(mn)$ time complexity ($m,n$ are the lengths of the involved two series). Maximizing the DTW speed is essential for many uses. In real-time health monitoring systems, for example, speed deficiency forces current systems to do coarse-grained sampling, leaving many sensed data unprocessed, causing risks of missing crucial symptoms early. For some diseases, it could be life threatening. For Septic Shock, for instance, every hour of delay in antibiotic treatment leads to an 8\% increase in mortality~\cite{Kumar2006}.

Many studies have been conducted to enhance the speed of DTW-based analytics~\cite{Rak+:TKDD2013,Tan+:SDM17,Yi+:VLDB2000,Kim+:ICDE2001,Lemire+:PR2009,Zhou+:ICDE2011}. 
The most commonly used is the use of lower bounds. Consider a common use of DTW, searching for the candidate series in a collection that is the most similar to a query series. We can avoid computing the DTW distance between the query and a candidate series if the lower bound of their DTW distance exceeds the best-so-far (i.e., the smallest DTW distance between the query and the already examined candidate series). Various lower bounding methods have been proposed~\cite{Yi+:VLDB2000,Kim+:ICDE2001,Lemire+:PR2009,Zhou+:ICDE2011}, with LB\_Keogh~\cite{Keogh+:VLDB2006} being the most popular choice. 

Most prior work on accelerating DTW has been, however, on univariate DTW; research on optimizing multivariate DTW has been scarce. A multidimensional or as we will refer to in the following, a multivariate time series is one in which each temporal point of the series carries multiple values. Multivariate time series data are ubiquitous in real-world dynamic systems such as health care and distributed sensor networks. For example, measurements taken from a weather station would be a multivariate time series with the dimensions being temperature, air pressure, wind speed, and so on.  There is a large demands for fast DTW on multivariate time series~\cite{Shokoohi+:DMKD17,Ridgely+:2009, Kela+:2006, Ko+:2005, Liu+:Mob2009, Peti+:TGRS12, Wang+:interspeech2013}.

The current multivariate DTW analysis is based on an algorithm named LB\_MV~\cite{Rath+:2003}. It was proposed in 2003 as a straightforward extension of the univariate lower bound-based DTW algorithms to multivariate data. There have been a few attempts to speed up multivariate DTW, but they are primarily concentrated on the reduction or transformations of the high dimension problem space~\cite{Li+:physica2015,Li+:elsevier2017,Hu+:ICDM2013,Gong+:Springer2015}, rather than optimizing the designs of DTW algorithms. 

The goal of this work is to come up with new algorithms that can significantly, consistently outperform the two-decade prevailing algorithm, whereby advancing the state of the art and better meeting the needs of temporal data analytics.







Our approach is creatively tightening the DTW lower bounds. The classic method, LB\_MV, gets a lower bound by efficiently computing a bounding envelope of a time series by getting the max and min on each dimension in a series of sliding windows. The envelope is loose, as Figure~\ref{fig:windowDTW}(d) shows (further explained in Section~\ref{sec:problem}). The key challenge in tightening the lower bounds is in keeping the time overhead sufficiently low while tightening the bounds as much as possible. It is especially challenging for multivariate data: As the DTW distance is affected by all dimensions, the computations of the lower bounds would need to take them all into consideration. 




Our solution is TC-DTW. TC-DTW addresses the problem by creatively introducing geometry and quantization into DTW bounds calculations. More specifically, TC-DTW introduces \textbf{\underline{T}riangle Inequality} and \textbf{Point \underline{C}lustering} into the algorithm design of multivariate \underline{DTW}. It features two novel ideas: (i) leveraging the overlapping windows of adjacent points in a time series to construct triangles and then using {\em Triangle Inequality} to compute tighter lower bounds with only scalar computations; (ii) employing {\em quantization-based point clustering} to efficiently reduce the size of bounding boxes in the calculations of lower bounds. These two ideas each lead to an improved algorithm for DTW lower bound calculations. Putting them together via a simple dataset-based adaptation, we show that the resulting TC-DTW consistently outperforms the classic algorithm LB\_MV, across datasets, data sizes, dimensions, temporal window sizes, and hardware.


Experiments on the 13 largest datasets in the online UCR collection~\cite{Bagnall+:arxiv2018} on two kinds of machines show that TC-DTW produces lower bounds that can help avoid as much as 98\% (60\% average) DTW distance calculations in DTW-based nearest neighbor findings, the most common usage of DTW. The speedups over the default windowed DTW (without lower bounds) are as much as 25$\times$ (8$\times$ on average), significantly higher than the speedups from the classic algorithm. Sensitivity studies on different dimensions, window sizes, and machines have all confirmed the consistently better performance of TC-DTW at no loss of any result quality, which suggests the potential of TC-DTW in serving as a more efficient drop-in replacement of existing multivariate DTW algorithms. The code has been released on github\footnote{https://github.com/DanielSongShen/MultDTW}. To our best knowledge, TC-DTW is the first known multivariate DTW algorithm that consistently outperforms the classic algorithm LB\_MV~\cite{Rath+:2003} with no loss of the result precision. 

In the rest of this paper, we first describe the background in Section~\ref{sec:background}, provide a formal problem statement~\ref{sec:problem}, and then present the integration of Triangle Inequality and Point Clustering into DTW respectively in Sections~\ref{sec:triangle} and~\ref{sec:clustering}. After that, we explain the adaptive deployment of two ideas, and their combined result TC-DTW in Section~\ref{sec:adaptive}. We report experimental results in Section~\ref{sec:experiments}, discuss some other issues in Section~\ref{sec:discussions} and related work in Section~\ref{sec:related}, and finally conclude the paper in Section~\ref{sec:conclusions}.

\section{Background}
\label{sec:background}
In this section, we will review the basic DTW algorithm as well as two common optimizations, {\em window DTW} and {\em lower bound filtering}. Then, we will describe the prior solutions to multivariate DTW and their limitations.

\mypar{Dynamic Time Warping (DTW)}
Let $Q = <q_1, q_2, ... q_m>$ be a {\em query} time series and $C = <c_1, c_2, ... c_n>$ be a {\em candidate} time series we want to compare it to. As Figure~\ref{fig:windowDTW} (a) illustrates, DTW tries to find the best alignment between $Q$ and $C$ such that the accumulated difference between the mapping points is minimum. In the mapping, it is possible that multiple points in one series map to one point in the other series. DTW finds such a mapping by building a cost matrix $C$ between each pair of points in $Q$ and $C$ where each element $d(i, j)$ of the matrix is the cost of aligning $q_i$ with $c_j$, as illustrated by the meshed box in Figure~\ref{fig:windowDTW}(b). DTW then uses dynamic programming, shown as the following recursive formula, to minimize the sum of the elements on a path from the bottom left to the top right of the cost matrix. The path is known as the {\em warping path}; at each element of the cost matrix, the path goes right, up, or up right.
\begin{equation}
    DTW(i, j) = d(i, j) +
    \min\begin{cases}         
        DTW(i-1, j) \\
        DTW(i, j-1) \\
        DTW(i-1, j-1) \\
    \end{cases}
\end{equation}

\mypar{Window DTW}
The basic DTW has a complexity O$(mn)$, where $m$ and $n$ are the lengths of the two sequences. A common approach to reducing the computations is to use a window to constrain the {\em warping path}. This restrains the distance in the time axis that a point $q_i$ can be mapped to in $C$. There are several versions of {\em window} DTW, the most popular of which, also the one we will be using in this paper, is the Sakoe-Chiba Band~\cite{SakoeBand}. 
As shown in Figure~\ref{fig:windowDTW} (b), the Sakoe-Chiba Band assumes the best path through the matrix will not stray far from the middle diagonal. Let two points aligning in C and Q be their $i$th and $j$th point respectively. The window constraint states that the following must hold: $j-W\le i\le j+W$, where $W$ is called the {\bf warping window size}. 
Such a constraint is used in almost all applications of DTW, in both single and multiple variate cases.

\mypar{DTW-NN and Univariate Lower Bound Filtering}
A common use of DTW is in finding a reference sequence that is most similar to a given query sequence---which is called {\em DTW-NN (or DTW Nearest Neighbor) problem}. It is, for instance, a key step in DTW-based time series clustering. As many DTW distances need to be computed when there are many reference sequences, even with the window constraints, the process can still take a long time. Many studies have tried to improve the speed. Lower bound filtering is one of the most widely used optimizations, designed for saving unnecessary computations and hence getting large speedups in this setting. The basic idea is that if the lower bound of the DTW distance between Q and a reference C is already greater than the best-so-far (i.e., the smallest distance from Q to the references examined so far), the calculations of the DTW distance between Q and C can be simply skipped as it is impossible for C to be the nearest neighbor of Q.

For univariate DTW, researchers have proposed a number of ways to compute DTW lower bounds (as listed in Section~\ref{sec:related}). The most commonly used is {\em LB\_Keogh} bound~\cite{Keogh+:VLDB2006}. Given two univariate time series $Q$ and $C$, of lengths $m$ and $n$ respectively, the LB\_Keogh lower Bound $LB(Q, C)$ is calculated in two steps. 

1) As shown in Figure~\ref{fig:windowDTW} (c), a {\em bounding envelope} is created around series $Q$ (red line) by constructing two time series $U$ and $L$ (for {\em upper} and {\em lower}, represented by two dotted lines) such that $Q$ must lie in the between. $U$ and $L$ are constructed as: 
\[
u_i = \max(q_{i-W} : q_{i+W})\;\;\;\;\;\;
l_i = \min(q_{i-W} : q_{i+W})
\]
 

2) Once the {\em bounding envelope} has been created, $LB(Q, C)$ can be simply computed from the parts of $C$ outside the envelope. Formally, it is calculated as the squared sum of the distances from every part of the candidate sequence $C$ outside the bounding envelope to the nearest orthogonal edge of the bounding envelope, illustrated as the shaded area in Figure~\ref{fig:windowDTW}(c).


\mypar{Multivariate DTW}
Multivariate DTW deals with multi-dimensional time series (MDT), where each point in a series is a point in a $D$-dimensional ($D>1$) space. We can represent a time series $Q$ as $q_1$, $\ldots$ $q_n$ and each $q_i$ is a vector in $R^D$, $(q_{i, 1}$, $\ldots$, $q_{i, D})$. 

Multivariate DTW measures the dynamic warping distance between two MDTs. There are two ways to define multivariate DTW distance. 

One is {\em independent DTW}, calculated as the cumulative distances of all dimensions {\em independently} measured under univariate DTW, expressed as $DTW_I(Q,C) = \sum_{i=1,D} DTW(Q_i, C_i)$, where $Q_i,C_i$ are univariate time series equaling the $i$th dimension of $Q$ and $C$ respectively. The other is {\em dependent DTW}, calculated in a similar way to the standard DTW for single-dimensional time series, except that the distance between two points on $Q$ and $C$ is the Euclidean distances of two $D$-dimensional vectors. Both kinds of DTWs have been used. A comprehensive study~\cite{Shokoohi+:DMKD17} has concluded that both kinds are valuable for practical applications. 

As independent DTW is the sum of multiple univariate DTWs, it can directly benefit from optimizations designed for univariate DTWs. Dependent DTW is different; the multi-dimensional distance calculations are not only more expensive than univariate distance, but also offer a different problem setting for optimizations. This paper focuses on optimizations of dependent DTW, particularly the optimizations that are tailored to its multi-dimensional nature. Without further notice, in the rest of this paper, multivariate DTW refers to dependent DTW. (For the completeness of the discussion, we note that there is another kind of multivariate DTW, which aligns points not on the time dimension, but on a multi-dimensional space~\cite{Cai+:NIPS2019}, such as matching pixels across two 2-D images. It is beyond the scope of this work.)




\section{Problem Statement and Design Considerations}
\label{sec:problem}

This section starts with a formal definition of the optimal lower bounding problem of multivariate DTW and the tradeoffs to consider in the solution design. It prepares the foundation for the follow-up presentations of the set of new algorithm designs. 

The optimal lower bounding problem is to come up with a way to calculate the lower bound of the DTW of two multivariate series such that when it is applied to DTW-NN (DTW for Nearest Neighbor introduced in Sec~\ref{sec:background}), the overall time is minimized. Formally, it is defined as follows.

\begin{definition}
{\bf Optimal Lower Bounding Problem for Multivariate DTW in DTW-NN.} Let $\mathbb{Q}$ and $\mathbb{C}$ be two sets of multivariate time series. The problem is to find $\mathbb{LB}^\ast = \{LB_{i,j}^\ast | i=1,\ldots,|\mathbb{Q}|$, $j=1,\ldots,|\mathbb{C}|$\} such that:
\begin{gather}
   \forall Q_i\in \mathbb{Q}, C_j\in \mathbb{C}, \;\;\;\;\; LB_{i,j}^\ast\le DTW(Q_i,C_j)\\
   \mathbb{LB}^\ast = \argmin_{\mathbb{LB}} T_{\mathbb{Q},\mathbb{C}}(\mathbb{LB}) + T_{\mathbb{LB}}(\mathbb{Q},\mathbb{C})\label{eq:sumtime}
\end{gather}

\noindent where, $DTW(Q_i,C_j)$ is the DTW distance between $Q_i$ and $C_j$, $T_{\mathbb{Q},\mathbb{C}}(\mathbb{LB})$ is the time taken to compute all lower bounds in $\mathbb{LB}$ from $\mathbb{Q}$ and $\mathbb{C}$, and $T_{\mathbb{LB}}(\mathbb{C},\mathbb{Q})$ is the time taken to find the nearest neighbor of $Q_i$ ($i=1,\ldots,|\mathbb{Q}|$) in $\mathbb{C}$ while leveraging $\mathbb{LB}$. 
\end{definition}

Formula~\ref{eq:sumtime} in the problem definition indicates the fundamental tradeoff in lower bounding, that is, the tension between the time needed to compute the lower bounds and the time the lower bounds may save DTW-NN. The tighter the lower bound is, the more effective it can be in helping avoid unnecessary DTW distance calculations; but if it takes too much time to compute, the benefits could be outweighed by the overhead. 

\mypar{Algorithm LB\_MV}
LB\_MV is the classic algorithm for computing the lower bounds of multivariate DTW. Since it was proposed in 2003~\cite{Rath+:2003}, it has been the choice in almost all applications of multivariate DTW. It is a simple extension of the univariate method for computing the {\em LB\_Keogh} lower bound (which is explained in Section~\ref{sec:background}). Like in the univariate case, LB\_MV also first builds a bounding envelope for the query series. Each point on the top and bottom of the envelope, $u_i$ and $l_i$, are defined as follows:

\[
u_i = (u_{i,1}, u_{i,2}, \ldots, u_{i,D})\;\;\;\;\;
l_i = (l_{i,1}, l_{i,2}, \ldots, l_{i,D})
\]

\noindent where, $u_{i,p}$ and $l_{i,p}$ are the max and min of the points in a window (centered at $i$) on the query series Q on dimension $p$, that is:
\[
u_{i,p} = max(q_{i-W,p} : q_{i+W,p})\;\;\;\;\;
l_{i,p} = min(q_{i-W,p} : q_{i+W,p})
\]

Figure~\ref{fig:windowDTW}(d) illustrates the bounding envelope of a 2D series in an actual time series of Japanese vowel audio~\cite{UCRArchive2018}. 

With the envelope, the lower bound of the DTW distance between series Q and series C is calculated as ($n$ for the length of $C$)

\[
LB\_MV(Q,C) = \sqrt{\sum_{i=1}^{n} \sum_{p=1}^{D} 
\begin{cases}
\begin{array}{ll}
(c_{i,p} - u_{i,p})^2 & if c_{i,p}>u_{i,p} \\
(c_{i,p} - l_{i,p})^2 & if c_{i,p}<l_{i,p} \\
0     & otherwise
\end{array}
\end{cases}
}
\]

This straightforward extension has been adopted widely in the applications of multivariate DTW. It computes the bounds fast, but is limited in effectiveness. As simple $max$ and $min$ are taken as the bounds in every dimension, the envelope is subject to the stretching effects on all dimensions, resulting in the loose bounding envelope as seen in Figure~\ref{fig:windowDTW}(d). (Experiments in Section~\ref{sec:experiments} confirm the limitation quantitatively.)

\begin{figure*}
    \centering
    \begin{tabularx}{0.9\textwidth}{c c c}
    \multicolumn{1}{m{4cm}}{\includegraphics[width=.3\textwidth]{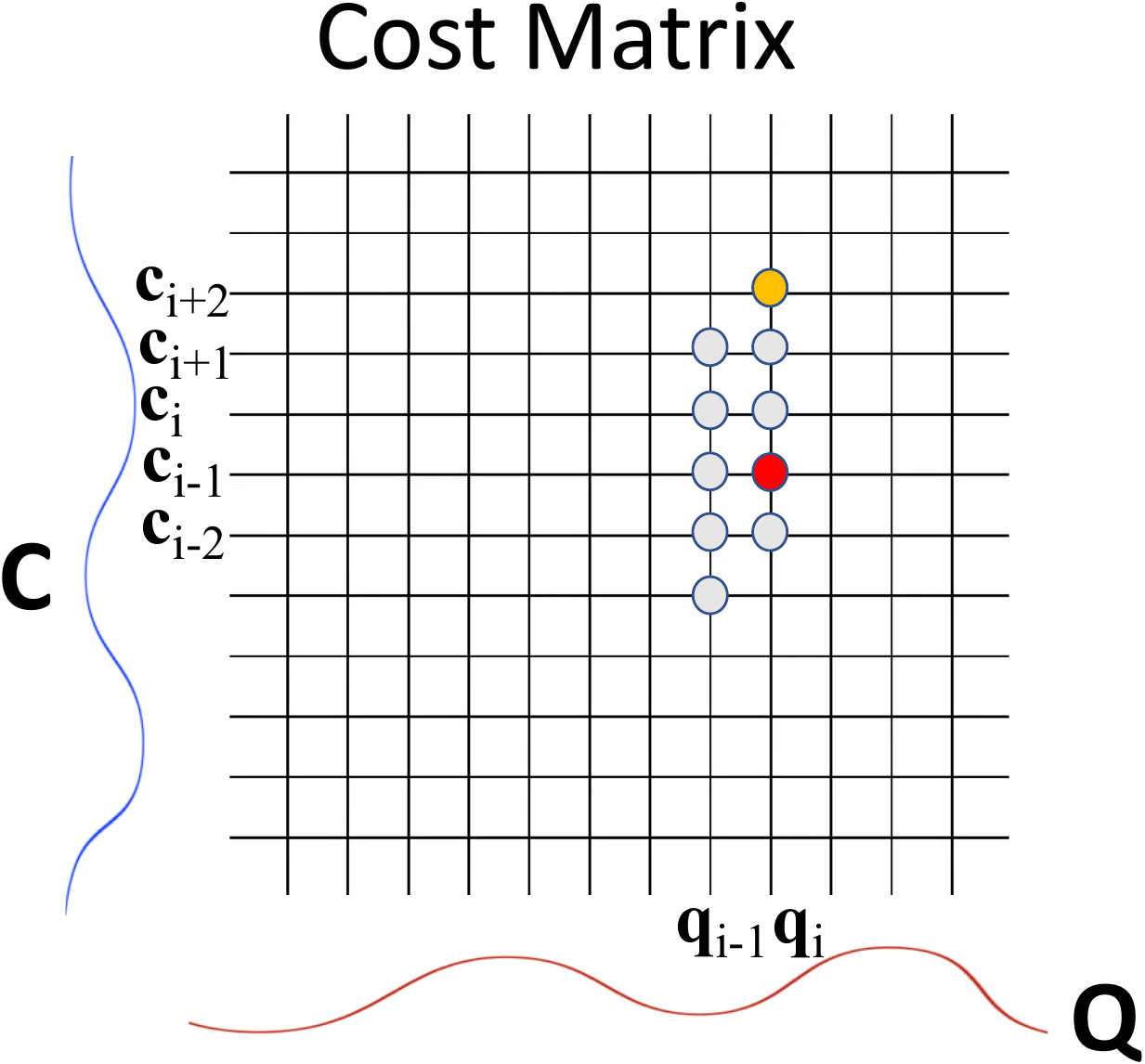}} & 
    $\;\;\;\;\;\;\;$ \includegraphics[width=.25\textwidth]{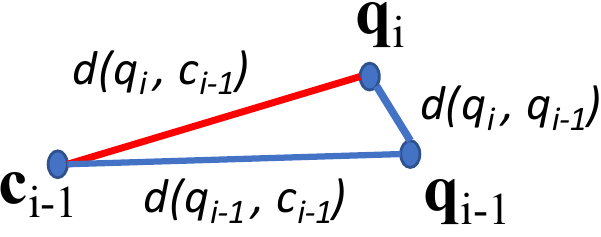} &
    $\;\;\;$ \includegraphics[width=.25\textwidth]{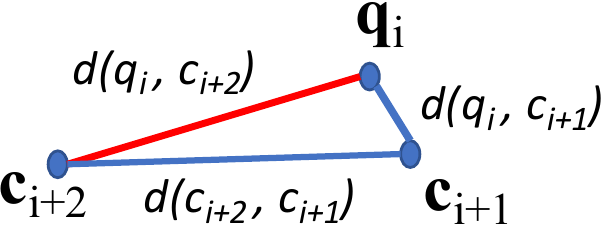}\\
    \multicolumn{1}{c}{(a)} & (b) & (c)
    \end{tabularx}
    \caption{(a) Cost matrix with window size equaling five. In LB\_AD, all the distances are computed between a candidate point and the query points within its window, the minimum of which is taken as the lower bound at the candidate, and the summation of these lower bounds across all candidate points is taken as the lower bound of $DTW(Q, C)$. (b) Illustration of LB\_TI at point $q_i$ and $c_{i-1}$. (c) Illustration of LB\_TI at the top boundary of a window, which is $c_{i+2}$ for query point $q_i$.}
    \label{fig:ad}
    \label{bdtw}
\end{figure*}

\mypar{Algorithm LB\_AD.} To facilitate the following discussions, we present a simple alternative method to LB\_MV illustrated in Figure~\ref{fig:ad}(a). In the graph, $Q$ and $C$ are a candidate series and a query series respectively. For a point $q_i$ on $Q$, this method computes the distances between $q_i$ and every candidate point in the relevant window, that is, every $c_j$ on $C$ ($i-W\le j\le i+W$), and selects the smallest distance as the point lower bound at $_i$. It then sums up all these point lower bounds of $C$ to get the lower bound of $DTW(C, Q)$. The soundness of the method is easy to see if we notice that the point that $q_i$ maps to in the DTW path must be one of the $c_j$s on $C$ ($i-W\le j\le i+W$). 

The lower bounds obtained by this method can be much tighter than LB\_MV, but the computation time to get the lower bounds is comparable to the time needed to compute the actual DTW distance. We call this method LB\_AD (AD for all distances).

The algorithms that we will introduce next can be regarded as optimizations to $LB\_AD$. They either leverage Triangle Inequality or Point Clustering to conservatively approximate the lower bound LB\_AD while substantially reducing the needed computations.

\section{Triangle Lower Bounding}
\label{sec:algorithms}
\label{sec:triangle}

The algorithms to be presented in this section center around the basic triangle inequality theorem in geometry. 

\begin{theorem}
{\bf Triangle Inequality:} for any triangle, the sum of the lengths of any two sides must be no smaller than the length of the remaining side, denoted as $d(a, b)\le d(a,c) + d(b,c)$, where d(x,y) is the distance between two points $x$ and $y$. 
\end{theorem} 

Triangle inequality holds on metrics, such as Euclidean distance. Although it does not directly apply to DTW distance as it is not a metric, an important insight we note is that it still holds for {\em point distances} in commonly seen time series, and leveraging it can avoid many unnecessary distance calculations and hence help efficient derivation of DTW lower bounds. We next explain how we turn that insight into more efficient DTW algorithms, which we call {\em triangle DTW}.


\subsection{Basic Triangle DTW (LB\_TI)}
\label{sec:bdtw}

Figure~\ref{bdtw}(b) illustrates the essential idea of the basic algorithm of triangle DTW, named LB\_TI (TI for Triangle Inequality). Instead of computing the distance between $q_i$ and every $c_j$ ($i-W\le j\le i+W$) as in LB\_AD, it uses triangle inequality to quickly estimate the lower bound of the distance $d(q_i, c_j)$. 

Consider an example when $j=i-1$, that is, the distance between $q_i$ and $c_{i-1}$ in the cost matrix in Figure~\ref{fig:ad}(a). These two points and $q_{i-1}$ together form a triangle, shown in Figure~\ref{fig:ad}(b). According to Triangle Inequality, we have

\begin{align}
    d(q_i, c_{i-1}) &\ge |d(q_{i-1}, c_{i-1}) - d(q_{i-1}, q_{i})|
\end{align}

So, $|d(q_{i-1}, c_{i-1}) - d(q_{i-1}, q_{i})|$ could be taken as the lower bound of $d(c_i, q_{i-1})$. If we do not know $d(q_{i-1}, c_{i-1})$, we can substitute it with its lower bound $L(q_{i-1}, c_{i-1})$ or upper bound $U(q_{i-1}, c_{i-1})$ as follows:

\begin{align}
    d(q_i, c_{i-1}) &\ge |d(q_{i-1}, c_{i-1}) - d(q_{i-1}, q_{i})| \nonumber\\
                    &= \max(d(q_{i-1}, c_{i-1}) - d(q_{i-1}, q_{i}), \;
                      d(q_{i-1}, q_{i}) - d(q_{i-1}, c_{i-1})) \nonumber\\
                    &\ge \max(L(q_{i-1}, c_{i-1}) - d(q_{i-1}, q_{i}), \;
                      d(q_{i-1}, q_{i}) - U(q_{i-1}, c_{i-1})) \label{ieq:bt_lb}
\end{align}

Similarly, via Triangle Inequality, we can get the upper bound of $d(q_i, c_{i-1})$ as follows:

\begin{align}
     d(q_i, c_{i-1}) &\le d(q_{i-1}, c_{i-1}) + d(q_{i-1}, q_{i}) \nonumber\\
                     & \le U(q_{i-1}, c_{i-1}) + d(q_{i-1}, q_{i}) \label{ieq:bt_ub}
\end{align}

Based on Equations~\ref{ieq:bt_lb} and~\ref{ieq:bt_ub}, we have the following recursive formula for the lower bound and upper bound of $d(q_i, c_{i-1})$

\begin{align}
    L(q_i,c_{i-1}) &= \max(L(q_{i-1}, c_{i-1}) - d(q_{i-1}, q_{i}),\;
    d(q_{i-1}, q_{i}) - U(q_{i-1}, c_{i-1})) \label{eq:bt_lb}\\
    U(q_i, c_{i-1}) &= U(q_{i-1}, c_{i-1}) + d(q_{i-1}, q_{i}) \label{eq:bt_ub}
\end{align}

In application of Equations~\ref{eq:bt_lb} and~\ref{eq:bt_ub}, $d(q_{i-1}, q_{i})$ will need to be computed, but as it is needed to be done only once for a query series $Q$ and can be reused for all candidates series, the overhead is marginal. $L(q_{i-1}, c_{i-1})$ and $U(q_{i-1}, c_{i-1})$ are already computed when the algorithm works on $q_{i-1}$. The true distances at $q_0$ can be computed and used as the lower bounds and upper bounds at $q_0$ to initiate the application of the formulae.

It is easy to see that the formulae hold if we replace $c_{i-1}$ with any other candidate points. What we care about are only the candidate points within the window of $q_i$, that is, $c_j  (j=i-W,i-W+1,\ldots,i+W)$. As points $c_{i-W}, c_{i-W+1}, \ldots, c_{i+W-1}$ also reside in the window of $q_{W-1}$, the lower bound and upper bound values on the right hand side (rhs) of Equations~\ref{eq:bt_lb} and~\ref{eq:bt_ub} should be available when the algorithm handles $q_i$. But $c_{i+W}$ is not, and hence needs a special treatment. 

Figure~\ref{fig:ad} (c) illustrates the special treatment of $c_{i+W}$ at $q_i$. It constructs a triangle with $c_{i+W}$, $q_i$, and $c_{i+W-1}$. The bounds can be hence computed as follows:

\begin{align}
    L(q_i,c_{i+W}) &= \max(L(q_{i}, c_{i+W-1}) - d(c_{i+W-1}, c_{i+W}),
    \; d(c_{i+W-1}, c_{i+W}) - U(q_{i}, c_{i+W-1})) \label{eq:bt_lb_ir}\\
    U(q_i, c_{i+W}) &= U(q_{i}, c_{i+W-1}) + d(c_{i+W-1}, c_{i+W}) \label{eq:bt_ub_ir}
\end{align}

\noindent where, $L(q_i, c_{i+W-1})$ and $U(q_i, c_{i+W-1})$ are the results at $q_i$ and $c_{i+W-1}$, and $d(c_{i+W-1}, c_{i+W})$ needs to be computed ahead of time; but as it can be reused for many query series, the overhead is marginal. 

\begin{figure}
    \centering
    \includegraphics[width=.7\textwidth]{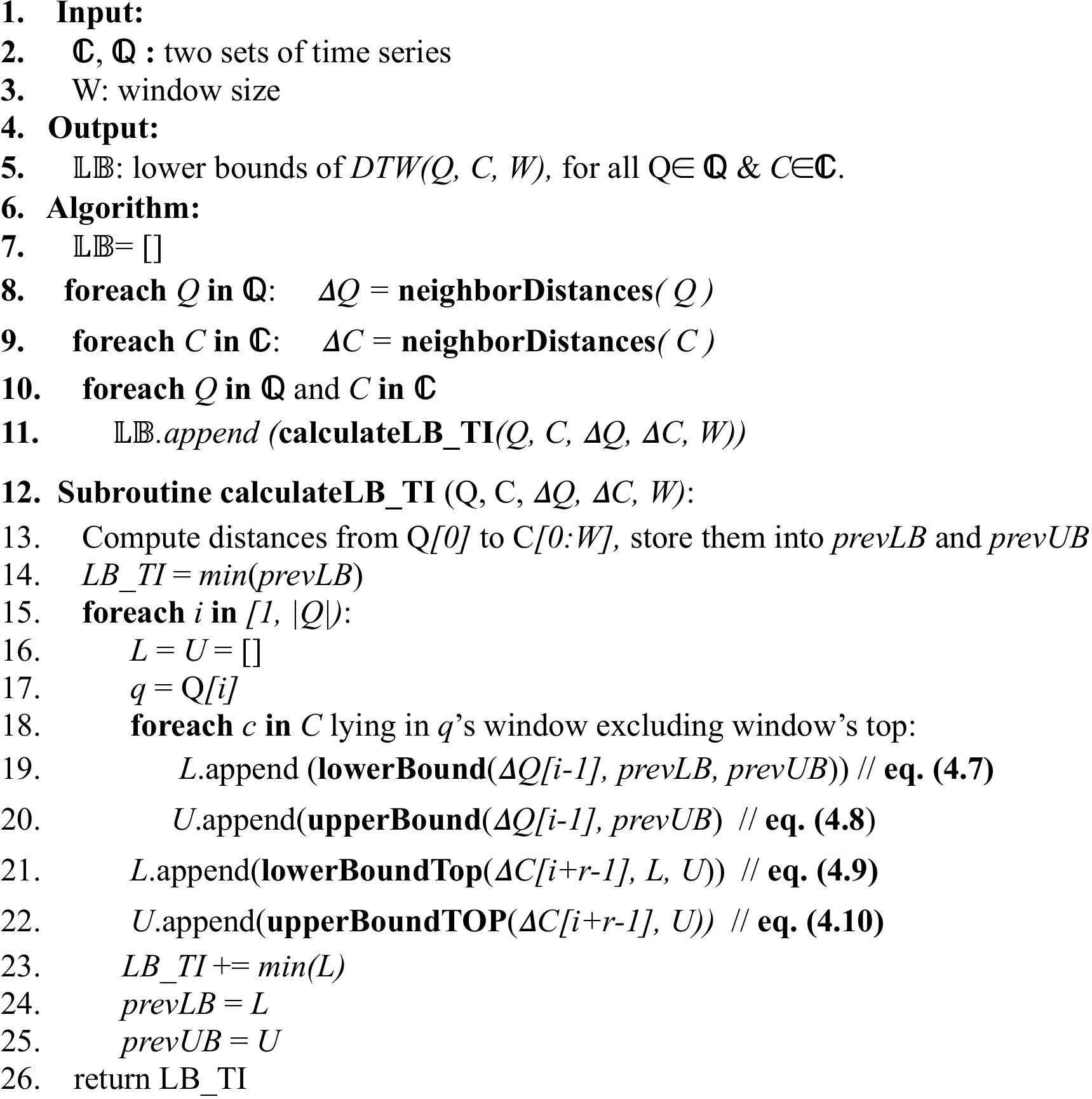}
    \caption{Algorithm of Basic Triangle DTW, $LB\_TI$.}
    \label{fig:bt_alg}
\end{figure}

\mypar{Full algorithm.} Figure~\ref{fig:bt_alg} outlines the entire algorithm. It first calculates the distances between every two adjacent points on each query and candidate series, then goes through each query-candidate series pair to compute the lower bound of their DTW. In the process, it applies Equations~\ref{eq:bt_lb}, \ref{eq:bt_ub}, \ref{eq:bt_lb_ir}, and \ref{eq:bt_ub_ir} to compute the lower bounds and upper bounds at each relevant query-candidate point pair; the minimum of the lower bounds of a candidate point is taken as its overall lower bound; the summation of all these overall lower bounds across all candidate points is taken as the lower bound of the DTW of the query-candidate series pair. 

\mypar{Characteristics.} Compared to $LB\_AD$, this algorithm saves computations by replacing vector (distance) computations with scalar (bound) computations. It replaces pair-wise distance computations in the distance matrix with lower bound and upper bound calculations. 
This basic Triangle DTW algorithm favors problems in which the time series are obtained through dense sampling such that the distance between two adjacent points on a time series is small: small $d(c_i, c_{i-1})$ and $d(q_{i+r}, q_{i+W-1})$ in Figure~\ref{fig:ad}(b,c) lead to tighter bounds. The recursive nature of Equations~\ref{eq:bt_lb} and~\ref{eq:bt_ub} entails that as the algorithm moves through the points on a candidate series, the bounds get looser and looser. We call this phenomenon the {\em diminishing tightness} problem. Some variations of the algorithm may mitigate the problem. We next describe several example variations.

\subsection{Variations of LB\_TI with Extra Distances}
\label{sec:extraDistance}

The variations in this section all try to tighten the bounds of the basic LB\_TI algorithm by adding some extra distance calculations. 

\mypar{(1) Triangle DTW with Top (LB\_TI\_TOP).} The differences of this variation from the basic LB\_TI is that it computes the true Euclidean distance between $q_i$ and the top boundary candidate point of its window, rather than estimates it with the ($q_i$, $c_{i+W}$, $c_{i+W-1}$) triangle (Equations~\ref{eq:bt_lb_ir} and~\ref{eq:bt_ub_ir}). Note that as a side effect, it forgoes the need for computing the distances between adjacent candidate points. 





\mypar{(2) Periodical Triangle DTW (LB\_TI$P$).} The difference of this variation from the basic LB\_TI is that it periodically tightens the bounds by computing the true candidate-query point distances at some candidate points, $q_i$ ($mod(i,P)==0)$, where $P$ is a positive integer, called the {\em TIP period length}. These distances are then used as the lower and upper bounds of $q_i$ in the follow-up bound calculations. The smaller $P$ is, the tighter the bounds could be, and the more computations it incurs. When $P$ equals the query series length $|Q|$, the algorithm is identical to the Basic Triangle algorithm LB\_TI. 


\mypar{(3) Periodical Triangle DTW with Top (LB\_TI$P$\_TOP).} It is a simple combination of LB\_TI$P$ and LB\_TOP, that is, besides periodically computing the true candidate-query point Euclidean distances, it also computes the true distances from every query point to the top boundary of its candidate window. 

Among these variations, our experiments show that variation three (LB\_TI$P$\_TOP) performs the best overall, striking an overall best tradeoff between the tightness of the bounds and the runtime overhead. In the following discussion, without further notice, we will use LB\_TI to refer to this variation. 

\section{Clustering DTW (LB\_PC)}
\label{sec:clustering}

The second approach we present tightens the lower bounds by exploiting the structure or distributions of data in time series through point clustering (LB\_PC). 

\begin{figure}
    \centering
     \begin{tabular}{ccc}
\includegraphics[width=.3\textwidth]{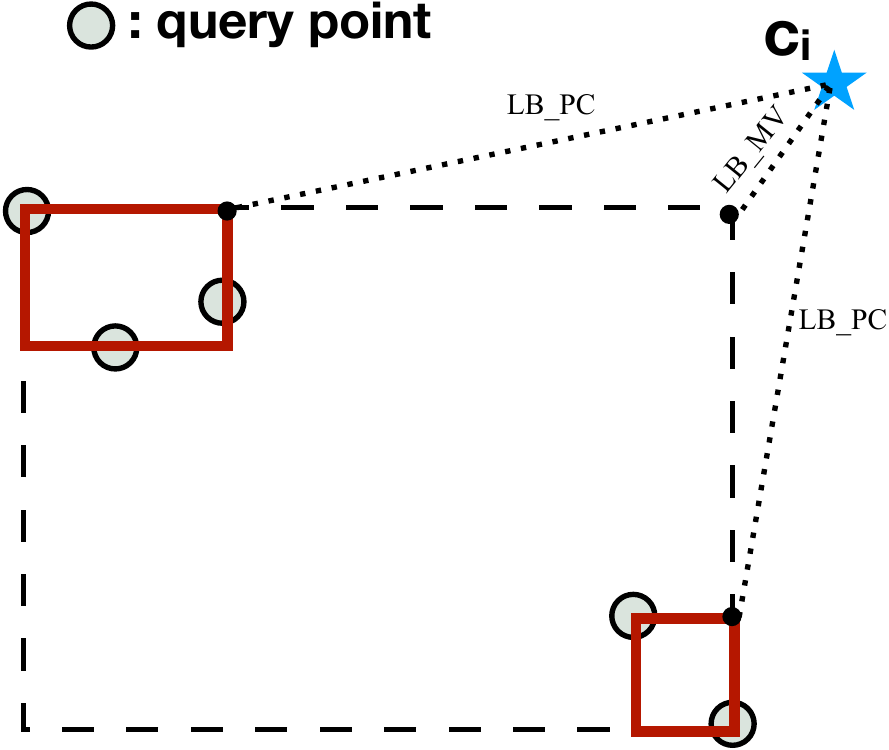} & $\;\;\;\;\;\;$ &
    \includegraphics[width=.5\textwidth]{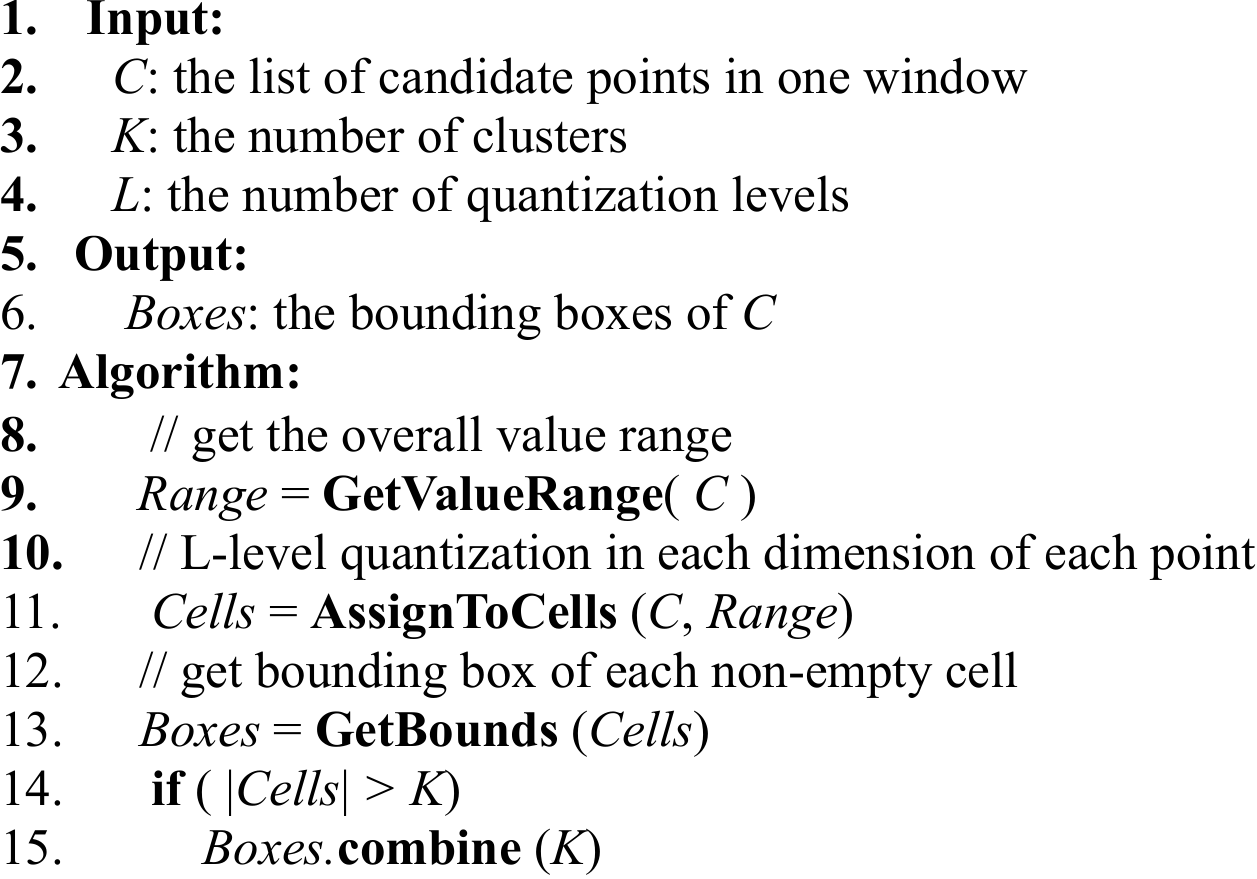}\\
(a) Insight behind LB\_PC & & (b) Algorithm in LB\_PC setup stage 
\end{tabular}
    \caption{(a) The insight behind LB\_PC. The five query points correspond to $q_j$ (j=i-2, i-1, i, i+1, i+2), the five points in the window of $c_i$. Clustering the query points allows the algorithm to better bound the query points and hence compute a lower bound much tighter than LB\_MV gets. (b) Algorithm used in the setup stage of LB\_PC for identifying the bounding boxes of candidate series.}
    \label{fig:point}
\end{figure}

Figure~\ref{fig:point}(a) illustrates the basic insight behind the design of LB\_PC. 
For the purpose of explanation, we assume that each point in the series of interest is a 2-D point, and the five query points, $q_{i-2}$ to $q_{i+2}$, are located in two areas as shown in Figure~\ref{fig:point}(a). An envelope used in LB\_MV is essentially a bounding box as shown by the dash lines in Figure~\ref{fig:point}(a), and the lower bound computed by LB\_MV is essentially the distance from $c_i$ to the corner closest to it which is corner $X$ in Figure~\ref{fig:point}(a). 

The bounds would be a lot tighter if we instead first cluster the five query points into two groups (the two solid-lined boxes in Figure~\ref{fig:point}(a)), apply LB\_MV to each of the groups to estimate the minimum distance between $c_i$ and each of the groups, and then take the minimum of those distances ($c_i$ to the two corners of the two solid-lined boxes in Figure~\ref{fig:point}(a)) as the lower bound between $c_i$ and the five query points. Our illustration uses 2-D space, but the insight obviously also applies to spaces of higher dimensions.



LB\_PC is designed on this insight. The challenge is to minimize the time overhead as clustering takes time. LB\_PC addresses it by using {\em adaptive quantization} as the clustering method, assisted by {\em uniform window grouping}. 

\mypar{Quantization-based clustering.} 

Figure~\ref{fig:point}(b) outlines the algorithm of quantization-based clustering. For some given points to cluster, it first gets the overall value range in each dimension. It then regards the value range as a grid with $L^D$ cells (for $D$ dimensions)---each dimension's value range is evenly divided into $L$ segments. The cell that a point belongs to can be then directly computed by dividing its value by the length of a cell in each dimension. Each non-empty cell is regarded as a cluster, and the boundaries of the cell define a bounding box. 

To control the number of clusters (e.g., no more than a certain number $K$), if the number of non-empty cells exceeds $K$, the last (in increasing lexicographic order of the cells) extra boxes are combined. In addition, we find that it is beneficial to add a limit on the smallest cell length in the quantization. When, for instance, all points in a window have very similar values in one dimension, separating the tiny range into multiple even smaller cells adds no benefits but overhead. In our experiments, we set the limit to 0.00001 of the normalized range of values. 

This clustering algorithm is applied by LB\_PC on the query points of each window. 
In essence, when a query comes, the LB\_PC algorithm works as LB\_MV does, except that it uses the bounding boxes (produced by the clustering algorithm) rather than the simple envelope. The value of $K$ and the quantization level $L$ determine the tradeoff between the tightness of the lower bounds and the lower bounds computation overhead. Its appropriate value can be empirically selected for a given dataset as other DTW parameters are. 

\mypar{Uniform window grouping.} 

To further reduce the cost, we introduce an optimization named {\em uniform window grouping}. It computes the bounding boxes for each {\em expanded window}. An expanded window is the result of merging $w$ consecutive windows. Take Figure~\ref{fig:ad} (a) as an example. In the original example, each window covers 5 points. The expanded window with $w=3$ would include 7 points \{$q_{i-2}$, $q_{i-1}$, $q_i$, $q_{i+1}$, $q_{i+2}$, $q_{i+3}$, $q_{i+4}$\}. The shifting stride becomes $w$ accordingly. The next expanded window would include points \{$q_{i}$, $q_{i+1}$, $q_{i+2}$, $q_{i+3}$, $q_{i+4}$, $q_{i+5}$, $q_{i+6}$\}, and so on. This method reduces the number of bounding boxes to compute and save and hence both the time and space cost by a factor of $w$. We call $w$ the {\em window expansion factor}.

Note that {\em expanded windows} are used in only the computations of bounding boxes; the DTW lower bound computations still use the original windows. In the DTW lower bound computation, the bounding boxes of an expanded window are used as the bounding boxes of the original windows that compose that expanded window. It may cause looser bounding boxes and hence looser lower DTW bounds, but experiments (Section~\ref{sec:experiments}) show that the impact is small. 

It is worth mentioning that an alternative grouping method we considered is non-uniform grouping, which groups adjacent windows if and only if their bounding boxes are the same. Experiments show that although this method keeps the bounding boxes sizes unchanged, the space saved is modest (about 20-30\%) and it cannot save time cost. The uniform grouping is hence preferred.

\section{Selective Deployment, TC-DTW, and Computational Complexities}
\label{sec:adaptive}

\mypar{Selective Deployment}
Both of the two proposed methods can help tighten the lower bounds, but also introduce extra time overhead. Inspired by the cascading idea in univariate DTW~\cite{Rak+:TKDD2013}, we use selective deployment of these algorithms with the combination of LB\_MV. For a given query-candidate series pair, the method first applies LB\_MV to get a lower bound. It applies the advanced algorithms only if $e<LB\_MV/d_{best}<1$, where $d_{best}$ is the DTW distance between the query series and the most similar candidate series found so far, and $e$ is called {\em triggering threshold}, a value between 0 and 1. The reason for that design is that if $LB\_MV/d_{best}\ge 1$, as the lower bound is greater than the best-so-far, this candidate series is impossible to be the most similar candidate; there is obviously no need to apply other methods. If the lower bound is a lot smaller than the best-so-far (i.e., $LB\_MV\le e$), there is a good chance for this current candidate to be indeed better than the best-so-far. Applying LB\_TI or LB\_PC may produce a larger lower bound, which is however unlikely to exceed $D_{best}$; the DTW between the query and the candidate series would then have to be computed. In both of these two cases, applying LB\_TI or LB\_PC adds overhead but avoids no extra DTW computations. This selective deployment strategy can help deploy DTW only when it is likely to give benefits. Similar to other DTW parameters, the appropriate value of {\em triggering threshold} $e$ for a dataset can be determined during the setup stage. 

\mypar{TC-DTW} LB\_TI and LB\_PC offer two different approaches to improve LB\_MV. As the next section shows, both are effective, but the one that performs the best could vary on different datasets. To get the best of both worlds, a simple method is to use a small samples of series, in the setup stage, to try out both methods and settle on the better one. We use TC-DTW to reference the resulting choice.



\begin{table}
    \begin{center}
    \caption{Summary of the algorithms on computing lower bound of DTW. (Notations at bottom)}\label{tab:analytic}
    \small
    \begin{tabularx}{\textwidth}{|c|X|c|}\hline
    \textbf{Method} & \multicolumn{2}{|c|}{\textbf{Lower Bounds Calculation}}\\\cline{2-3}
           & \textbf{Operations} & \textbf{Time Complexity} \\\hline
    LB\_MV & \noindent\parbox[c]{\hsize}{Compute envelope of Q and distances to bounding envelopes (1 bounding box/point)} & $\mathcal{O}(D\ast n\ast M\ast N)$\\\hline
    LB\_AD & \noindent\parbox[c]{\hsize}{Compute distances to each candidate point} & $\mathcal{O}(W\ast D\ast n\ast M\ast N)$ \\\hline
    LB\_TI & \noindent\parbox[c]{\hsize}{Compute neighbor distances on Q, 1/P LB\_AD distances, TI bounds, and window top distances} & $\mathcal{O}(4+((2+W/P)\ast D)\ast n\ast M\ast N)$ \\\hline
    LB\_PC & \noindent\parbox[c]{\hsize}{Compute bounding boxes and distances to bounding boxes ($K$ boxes per $w$ points)} & $\mathcal{O}(K\ast D\ast n\ast M\ast N/w)$\\\hline
    \end{tabularx}\\
\end{center}

     \scriptsize
     n: series length (for explanation purpose, assume |C|=|Q|); $M:|\mathbb{C}|$; $N:|\mathbb{Q}|$; $W:$ window size; $D:$ dimension; $P:$ the TIP period length;\\ $K:$ \# clusters in LB\_PC; $w:$ window expansion factor.
\end{table}

\mypar{Computational Complexities} Table~\ref{tab:analytic} summarizes the various methods for computing the lower bounds of DTW between two collections of time series $\mathbb{C}$ and $\mathbb{Q}$. For simplicity, the table assumes the same length shared by query and candidate series. LB\_AD and LB\_MV lie at two ends of the spectrum of overhead and tightness, while the methods proposed in this work offer options in the between. They are equipped with knobs (e.g., the period length $P$ and the number of clusters $K$) which can be adjusted to strike the best tradeoff for a given problem.

\section{Experiments}
\label{sec:experiments}

To evaluate the efficacy of the proposed algorithms, we conduct a series of experiments and comparisons in DTW-based nearest neighbor findings, in which, the goal is to find the most similar (i.e., having the smallest DTW distance) candidate series for each query series. In this most common use of DTW, lower bounds are used to avoid computing the DTW distance between a query and a candidate if their DTW lower bound already exceeds the so-far-seen smallest DTW distance. The experiments are designed to answer the following major questions:
\begin{enumerate}
    \item How effective are the two new methods, LB\_TI and LB\_PC, in accelerating multivariate DTW in nearest neighbor finding?
    \item How effective are the two techniques when they are applied without prior preparations in the streaming scenario? What effects does the runtime overhead of the lower bound calculations impose on the end-to-end performance?
    \item How do the benefits change with window sizes or data dimensions? 
    \item Do they perform well on different machines?
\end{enumerate}


\subsection{Methodology}
\label{sec:methodology}

\mypar{Methods to compare.} We focus our comparisons on the following methods:

\begin{itemize}
    \item LB\_MV: the most commonly used method in multivariate DTW~\cite{Rath+:2003}.
    \item LB\_TI: our proposed triangle DTW (period length $P$ is set to five for all datasets). 
    \item LB\_PC: our proposed point clustering DTW with {\em uniform window grouping}. 
    \item TC-DTW: the combination of LB\_TI and LB\_PC. For a given dataset, it picks the better one of the two methods as described in Section~\ref{sec:adaptive}. 
\end{itemize}

Both LB\_TI and LB\_PC use the selective deployment scheme as described in Section~\ref{sec:adaptive}. The parameters used in the two methods are selected from the following ranges: \{0.05, 0.1, 0.2\} for the triggering threshold in LB\_TI, \{0.1, 0.5\} for the triggering threshold in LB\_PC, \{2,3\} for the quantization levels in LB\_PC. 
The selection, as part of the database configuration process, is through empirically experimenting with the parameters on a small set (23) of randomly chosen candidate series. Other parameters are the maximum number of clusters $K$ and the window expansion factor $w$ in LB\_PC, which are both set to 6 for all datasets.

In all the methods, early abandoning~\cite{Wang+:DMKD2013} is used: During the calculations of either lower bounds or DTW distances, as soon as the value exceeds the so-far-best, the calculation is abandoned. This optimization has been shown beneficial in prior uni-variate DTW studies~\cite{Wang+:DMKD2013}; our observations on multivariate DTW are also positive. 

\mypar{Datasets.} Our experiments used all the 13 largest multivariate datasets in the online UCR multivariate data collection~\cite{Bagnall+:arxiv2018}, which are the ones in the most need for accelerations. 
The numbers of pair-wise DTW distances in these datasets range from 40,656 to 525 million, as Table~\ref{tab:datasets} shows. These datasets cover a wide variety in domains, the length of a time series (\texttt{Length}), and the numbers of dimensions of data points (\texttt{Dim}). Values on the time series were normalized before being used,
and NA values were replaced with zeros. For each dataset, 30\% are used as queries, and the rest are used as candidates.

\begin{table*}
\centering
\caption{Datasets}\label{tab:datasets}
\small
\begin{tabular}{|llllllp{.3\textwidth}|}
\hline
 Dataset & \#DTW  & \#Points & \#Series & Dim & Length  & Description \\\hline   
Articularyword.  & 69,431&	82,800&	575&	9&	144&	Motion tracking of lips and mouth while speaking 12 different words              \\\hline
Charactertraj.      & 1,715,314&	520,156&	2858&	3&	182& Pen force and trajectory while writing various characters on a tablet            \\\hline
Ethanolconcen.       & 57619&	917,524&	524&	3&	1751& Raw spectra of water-and-ethanol solutions in whisky bottles\\\hline
Handwriting                & 210,000&	152,000&	1000&	3&	152& Motion of writing hand when writing 26 letters of the alphabet.                  \\\hline
Insectwingbeat             & 525,000,000&	1,500,000&	50000&	200&	22& Reconstruction of the sound of insects passing through a sensor\\\hline
Japanesevowels             & 86,016&	18,560&	640&	12&	29& Audio, recording of Japanese speakers saying "a" and   "e"                       \\\hline
Lsst                       & 5,093,681&	177,300&	4925&	6&	36& Simulated astronomical time-series data in preparation for observations from the LSST Telescope\\\hline
Pems-sf                    & 40,656&	63,360&	440&	963&	144& Occupancy rate of car lanes in San Francisco bay area                            \\\hline
Pendigits                  & 25,373,053&	87,936&	10992&	2&	8& Pen xy coordinates while writing digits 0-9 on a tablet                          \\\hline
Phonemespectra             & 9,337,067&	1,446,956&	6668&	11&	217& Segmented audio collected from Google Translate Audio files \\\hline
Selfregulationscp1         & 66024&	502,656&	561&	6&	896& cortical potentials signal \\\hline
Spokenarabicdigits         & 16,255,009&	818,214&	8798&	13&	93& Audio recording of Arabic digits\\\hline
Uwavegesturelib.        & 4,211,021&	1,410,570&	4478&	3&	315& XYZ coordinates from accelerometers while performing various simple   gestures. \\\hline
\end{tabular}
\end{table*}

\mypar{Hardware and measurement.} To consider the performance sensitivity to hardware, we have measured the performance of the methods on two kinds of machines. (1) AMD machine with 2GHz Opteron CPUs, 32GB DDR3 DRAM, and 120GB SSD; (2) Intel machine with 2.10GHz Skylake Silver CPUs, 96GB DDR4 2666 DRAM, and 240GB INTEL SDSC2KB240G7 SSD. We focus the discussions on the results produced on the AMD machine and report the results on the Intel machine later. The machines run Redhat Linux 4.8.5-11.

In the experiments, the methods are applied on the fly by computing the envelopes, neighboring point distances, and bounding boxes of the query series at runtime as the query arrives. All runtime overhead is counted in the performance report of all the methods. The reported times are the average of 10 repetitions. 

\mypar{Window Sizes and Data Dimensions}
One of the aspects in the experiment design is to decide what window sizes to use for the DTW calculations. Our decision is based on observations some earlier work has made on what window sizes make DTW-based classification give the highest classification accuracy; the results are listed as "learned\_w" on the UCR website~\cite{UCRArchive2018}. One option considered is to use 
certain percentages of a sequence length as the window size as some prior work does. But observations on the UCR list show that there is no strong correlations between sequence length and the best window size. In fact, a percentage range often used by prior work is 3-8\% of the sequence length, which covers only 10\% of the 128 datasets. On the other hand, the best window sizes of almost all (except two) datasets fall into the range of [0, 20] regardless of the sequence length. Although that list does not contain most of the multivariant datasets used in this study, the statistics offer hints on the range of proper window sizes. Based on those observations, we experiment with two window sizes, 10 and 20. 

Even though some of the datasets have points with a high dimension, in practice, practitioners typically first use some kind of dimension reduction to lower the dimensions to a small number~\cite{Hu+:ICDM2013} to avoid the so-called curse of dimensionality. In all the literature we have checked on the applications of multivariate DTW, the largest dimension used is less than 10. Hence, in our examination of the impact of dimensionality on the methods, we study three settings: $dim$=3,5,10.

\begin{figure}
    \centering
    \begin{tabular}{cc}
    \includegraphics[width=.4\textwidth]{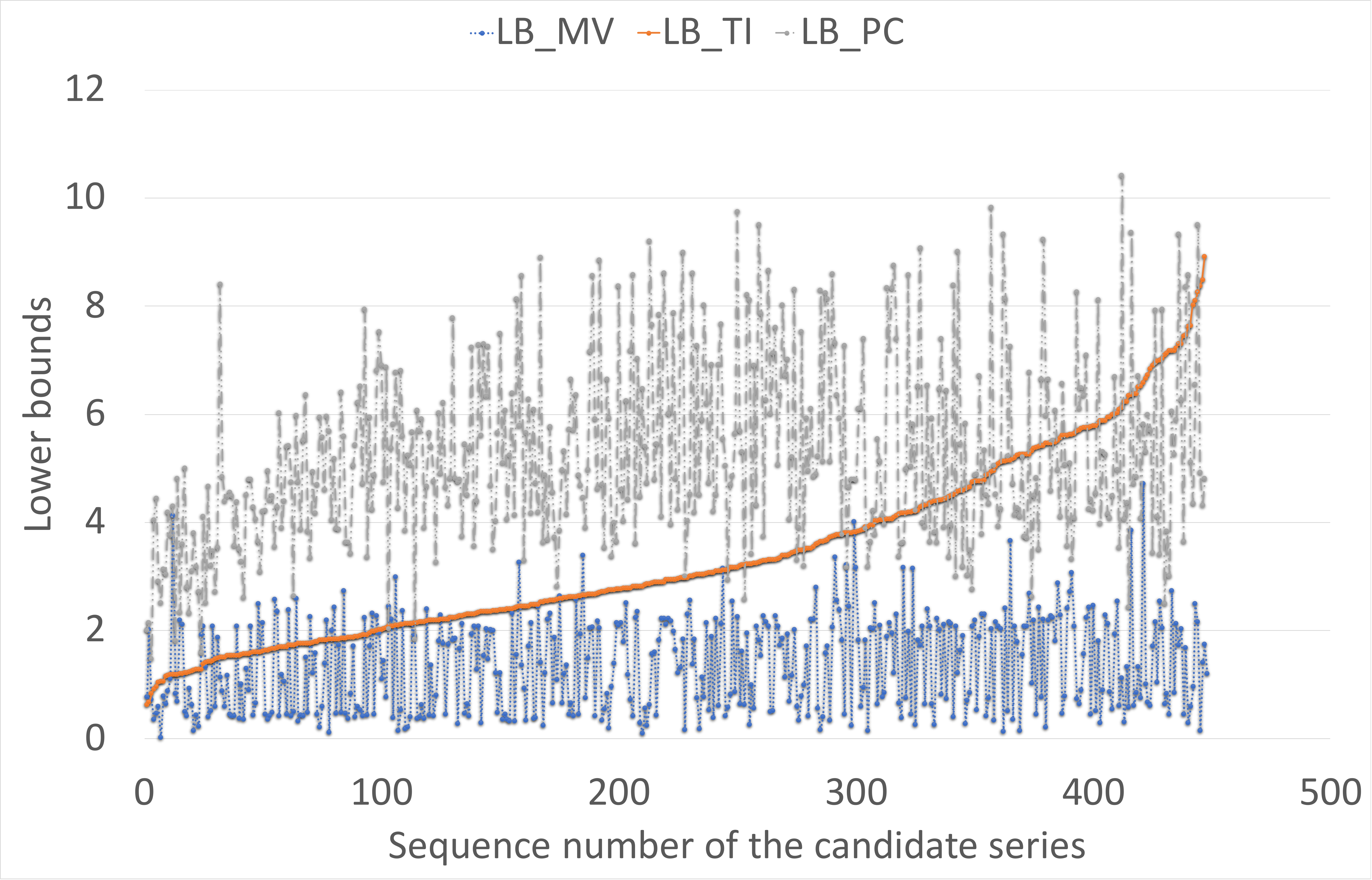}     &
    \includegraphics[width=.4\textwidth]{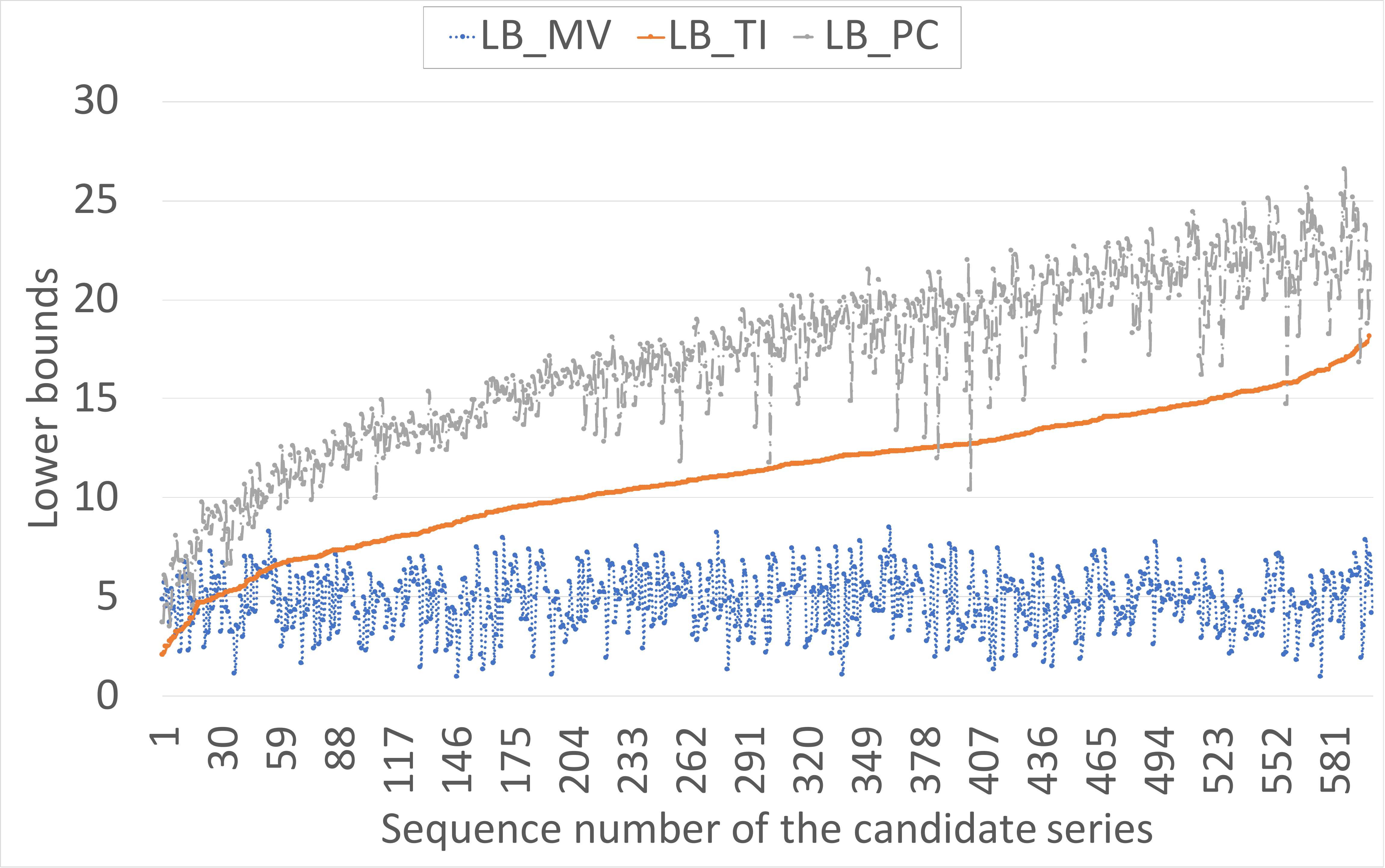}\\
    (a) Japanesevowels    & (b) Insectwingbeat
    \end{tabular}
    \caption{The lower bounds of the DTW distance between sequence 0 and other sequences in Japanesevowels and Insectwingbeat datasets. For legibility, the candidate series are sorted based on the LB\_TI bounds.}
    \label{fig:exampleBounds}
\end{figure}

\begin{table*}[]
\centering
\small
\caption{Comparisons of LB\_TI and LB\_PC over LB\_MV with all runtime overhead counted.}\label{tab:online}
\begin{tabular}{|l|*{4}{l}|*{4}{l}|}
\hline
\multicolumn{9}{|l|}{W=20*} \\\hline
                          & \multicolumn{4}{|l|}{Speedups}& \multicolumn{4}{l|}{Percentage of skipped DTWs calc.}
\\ \cline{2-9}
Dataset                   & LB\_MV      &TC-DTW & LB\_TI & LB\_PC & LB\_MV      &TC-DTW & LB\_TI & LB\_PC \\\hline
Articularyword. & 7.60& 10.51& 7.32& 10.51& 84\%& 91\%& 85\%& 91\% \\
Charactertraj.     & 16.98& 25.31& 22.11& 25.31& 89\%& 99\%& 92\%& 99\% \\
Ethanolconcen.      & 6.67& 11.08& 11.08& 7.06& 79\%& 87\%& 87\%& 80\% \\
Handwriting               & 3.41& 3.55& 3.55& 3.55& 36\%& 50\%& 42\%& 50\% \\
Insectwingbeat            & 6.03& 9.02& 7.66& 9.02& 56\%& 90\%& 71\%& 90\% \\
Japanesevowels            & 2.59& 2.91& 2.69& 2.91& 10\%& 40\%& 17\%& 40\% \\
Lsst                      & 1.68& 1.64& 1.64& 1.63& 7\%& 15\%& 7\%& 15\% \\
Pems-sf                   & 6.05& 7.33& 6.02& 7.33& 78\%& 85\%& 77\%& 85\% \\
Pendigits                 & 4.06& 4.13& 4.13& 3.89& 27\%& 48\%& 34\%& 48\% \\
Phonemespectra            & 1.95& 2.02& 1.92& 2.02& 12\%& 13\%& 13\%& 13\% \\
Selfregulationscp1        & 1.86& 1.92& 1.92& 1.92& 24\%& 29\%& 25\%& 29\% \\
Spokenarabicdigits        & 5.76& 7.71& 7.71& 7.28& 64\%& 83\%& 75\%& 83\% \\
Uwavegesturelib.       & 11.85& 17.63& 17.63& 13.89& 92\%& 95\%& 95\%& 94\% \\
\textbf{Average}          & 5.88& 8.06& 7.34& 7.41& 51\%& 63\%& 55\%& 63\% \\\hline
\multicolumn{9}{|l|}{W=10*} \\\hline
                          & \multicolumn{4}{|l|}{Speedups}& \multicolumn{4}{l|}{Percentage of skipped DTWs calc.}
\\ \cline{2-9}
Dataset                   & LB\_MV      &TC-DTW & LB\_TI & LB\_PC & LB\_MV      &TC-DTW & LB\_TI & LB\_PC \\\hline
Articularyword. & 11.24& 11.78& 10.68& 11.78& 96\%& 95\%& 94\%& 95\% \\
Charactertraj.     & 16.77& 18.52& 18.52& 17.78& 100\%& 100\%& 100\%& 99\% \\
Ethanolconcen.      & 13.29& 15.87& 15.87& 13.05& 97\%& 100\%& 100\%& 94\% \\
Handwriting               & 4.16& 4.69& 3.99& 4.69& 51\%& 72\%& 55\%& 72\% \\
Insectwingbeat            & 5.44& 10.91& 7.17& 10.91& 66\%& 94\%& 81\%& 94\% \\
Japanesevowels            & 2.55& 3.21& 2.88& 3.21& 30\%& 71\%& 45\%& 71\% \\
Lsst                      & 1.67& 2.06& 1.70& 2.06& 21\%& 39\%& 21\%& 39\% \\
Pems-sf                   & 7.28& 7.47& 6.78& 7.47& 86\%& 91\%& 85\%& 91\% \\
Pendigits                 & 4.08& 4.29& 4.29& 3.98& 27\%& 49\%& 34\%& 49\% \\
Phonemespectra            & 2.08& 2.16& 2.10& 2.16& 25\%& 27\%& 25\%& 27\% \\
Selfregulationscp1        & 2.10& 2.38& 2.18& 2.38& 41\%& 52\%& 41\%& 52\% \\
Spokenarabicdigits        & 8.19& 10.42& 10.07& 10.42& 86\%& 97\%& 90\%& 97\% \\
Uwavegesturelibrary       & 15.40& 16.17& 16.17& 15.23& 98\%& 100\%& 100\%& 98\% \\
\textbf{Average}                   & 7.25& 8.46& 7.88& 8.09& 63\%& 76\%& 67\%& 75\% \\\hline
\end{tabular}\\
\scriptsize
*: Actual window sizes are capped at the length of a time series.
\end{table*}

\begin{table}[]
\centering
\caption{Speedups ($\times$) in Various Dimensions on Datasets with More than 8 Dimensions. (W=20)}\label{tab:dims}
\begin{tabular}{|l|*{4}{l}|}
\hline
\multicolumn{5}{|l|}{Dim=3} \\\hline
Dataset                   & LB\_MV         & TC-DTW     & LB\_TI        & LB\_PC\\\hline
Articularyword.       & 7.67& 10.59& 7.49& 10.59\\
Insectwingbeat                 & 12.12& 15.24& 14.39& 15.24\\
Japanesevowels                  & 3.17& 3.53& 3.41& 3.53\\
Pems-sf                         & 5.99& 7.17& 5.98& 7.17\\
Phonemespectra                  & 1.97& 1.97& 1.97& 1.95\\
Spokenarabic.              & 8.41& 11.54& 11.01& 11.54\\
\textbf{Average}                & 6.55& 8.34& 7.38& 8.34\\\hline
\multicolumn{5}{|l|}{Dim=5} \\\hline
Articularyword.   & 6.86& 9.24& 6.86& 9.24\\
Insectwingbeat    & 6.13& 8.94& 8.04& 8.94\\
Japanesevowels    & 2.63& 2.97& 2.66& 2.97\\
Pems-sf           & 7.82& 8.58& 7.77& 8.58\\
Phonemespectra    & 1.93& 1.98& 1.98& 1.95\\
Spokenarabic.     & 5.55& 7.81& 7.81& 7.40\\
\textbf{Average}  & 5.15& 6.59& 5.85& 6.51\\\hline
\multicolumn{5}{|l|}{Dim=10} \\\hline
Articularyword.       & 5.80& 8.10& 5.87& 8.10\\\
Insectwingbeat                  & 2.55& 4.45& 3.21& 4.45\\
Japanesevowels                  & 2.12& 2.28& 2.20& 2.28\\
Pems-sf                         & 5.27& 5.94& 5.17& 5.94\\
Phonemespectra                  & 2.00& 1.97& 1.97& 1.87\\
Spokenarabic.              & 3.12& 4.09& 4.09& 3.93\\
\textbf{Average}        & 3.48& 4.47& 3.75& 4.43\\\hline
\end{tabular}
\end{table}

\begin{table}[]
\centering
\caption{Performance on Intel machine. (W=20)}\label{tab:intel}
\begin{tabular}{|l|*{4}{l}|}
\hline
\hline
                          & \multicolumn{4}{l|}{Speedup ($\times$)}                                  \\ \cline{2-5}
                          Dataset                   & LB\_MV        & TC-DTW        & LB\_TI        & LB\_PC      \\ \hline
                          Articularyword.           & 3.52& 4.88& 3.82& 4.88\\
                          Charactertr.              & 6.97& 9.99& 9.21& 9.99\\
                          Ethanolconc.      & 6.70& 11.16& 11.16& 7.02\\
                          Handwriting               & 2.83& 2.96& 2.96& 2.86\\
                          Insectwingbeat            & 9.09& 13.70& 11.65& 13.70\\
                          Japanesevowels            & 3.96& 4.48& 4.19& 4.48\\
                          Lsst                      & 2.37& 2.39& 2.26& 2.39\\
                          Pems-sf                   & 3.53& 3.86& 3.66& 3.86\\
                          Phonemespectra            & 2.44& 2.46& 2.46& 2.36\\
                          Pendigits                 & 6.00& 6.34& 6.17& 6.34\\
                          Selfregula.        & 1.82& 1.97& 1.88& 1.97\\
                          Spokenarabic.        & 5.62& 10.84& 10.45& 10.84\\
                          Uwavegesture.             & 5.06& 7.89& 7.20& 7.89\\
                          \textbf{Average}          & 4.61& 6.38& 5.93& 6.04\\\hline
\end{tabular}
\end{table}

\subsection{Experimental Results}

\mypar{Soundness and Sanity Checks} 
Because the optimizations used in all the methods keep the semantic of the original DTW distance, they do not affect the precision of the results. It is confirmed in our experiments: The optimized methods find the nearest neighbor for every query as the default method does, and return the correct DTW distance. In a sanity check on Nearest Neighbor-based classification, the classification results are identical among all the methods. We focus our following discussions on speed.

\mypar{Overall Speed} 
Table~\ref{tab:online} reports the results when the window size is set to 20 and 10; the actual window size for a dataset is capped at the length of a series. The first five dimensions are used; Table~\ref{tab:dims} reports results on other dimensions.

The speedup of a method X on a dataset is computed as $T(X)/T$(default), where $T$(default) is the time taken by the default windowed DTW algorithm (which uses no lower bounds) in finding the candidate series most similar to each query series. $T(X)$ is the time taken by method X which includes all the runtime overhead, such as the time of lower bounds calculations if method X uses lower bounds.

In Table~\ref{tab:online}, the "Speedups" columns report the speedups of the methods over the default algorithm. The "Skips" columns in Table~\ref{tab:online} report the percentage of the query-candidate DTW distances that those methods successfully avoid computing via their calculated lower bounds. 

From Table~\ref{tab:online}, We see that on average, LB\_MV gives 5.86$\times$ speedups on the datasets. The speedups come from avoiding the DTW calculations for 50\% of query-candidate pairs and also the early abandoning scheme. LB\_TI avoids 56\% DTW calculations, and gives 7.33$\times$ average speedup. LB\_PC avoids 64\% DTW calculations, achieving 7.42$\times$ average speedups, slightly higher than LB\_TI. As examples, Figure~\ref{fig:exampleBounds} shows the lowerbounds from the three methods on two representative datasets. Similar trends appear on other datasets. The larger lower bounds from LB\_TI and LB\_PC are the main reasons for the time savings. 

In most cases, LB\_PC avoids more DTW computations and gets a larger speedup than LB\_TI. There are four exceptions, \texttt{Ethanolconcentration, Pendigits, Spokenarabicdigits} and \texttt{Uwavegesturelibrary}, on which LB\_TI performs better than LB\_PC does. The results are influenced by the value distributions and changing frequencies of the datasets. Method TC\_DTW gets the best of both worlds through its dataset-adaptive selection, achieving an average 8.03$\times$ speedups. The speedups differ substantially across datasets. On \texttt{Lsst}, the avoidance rate is only 15\%, and the speedup is 1.67$\times$; on \texttt{Charactertrajectories}, the avoidance rate is as large as 98\%, and the speedup is 25.1$\times$. In general, if the series in a dataset do not have many differences, lower bounds are less effective in avoiding computations, and the speedups are less pronounced. 
Despite the many differences among the datasets, TC-DTW consistently outperforms the popular method LB\_MV on all datasets.  

The results also confirm the usefulness of early abandoning. For instance, even though LB\_MV avoids only 7\% DTW distance calculations on dataset \texttt{Lsst}, the speedup it gives is as much as 1.65$\times$, showing the benefits from the early abandoning. Nevertheless, the benefits from the improved DTW algorithms are essential. For instance, on dataset \texttt{Charactertrajectories}, the 25$\times$ speedups come primarily from the avoidance of 98\% of the DTW distance calculations. In fact, because early abandoning is adopted by all four methods shown in Table~\ref{tab:online}, the differences in their produced speedups are caused only by their algorithmic differences rather than early abandoning.

\begin{figure}
    \centering
    \includegraphics[width=.7\textwidth]{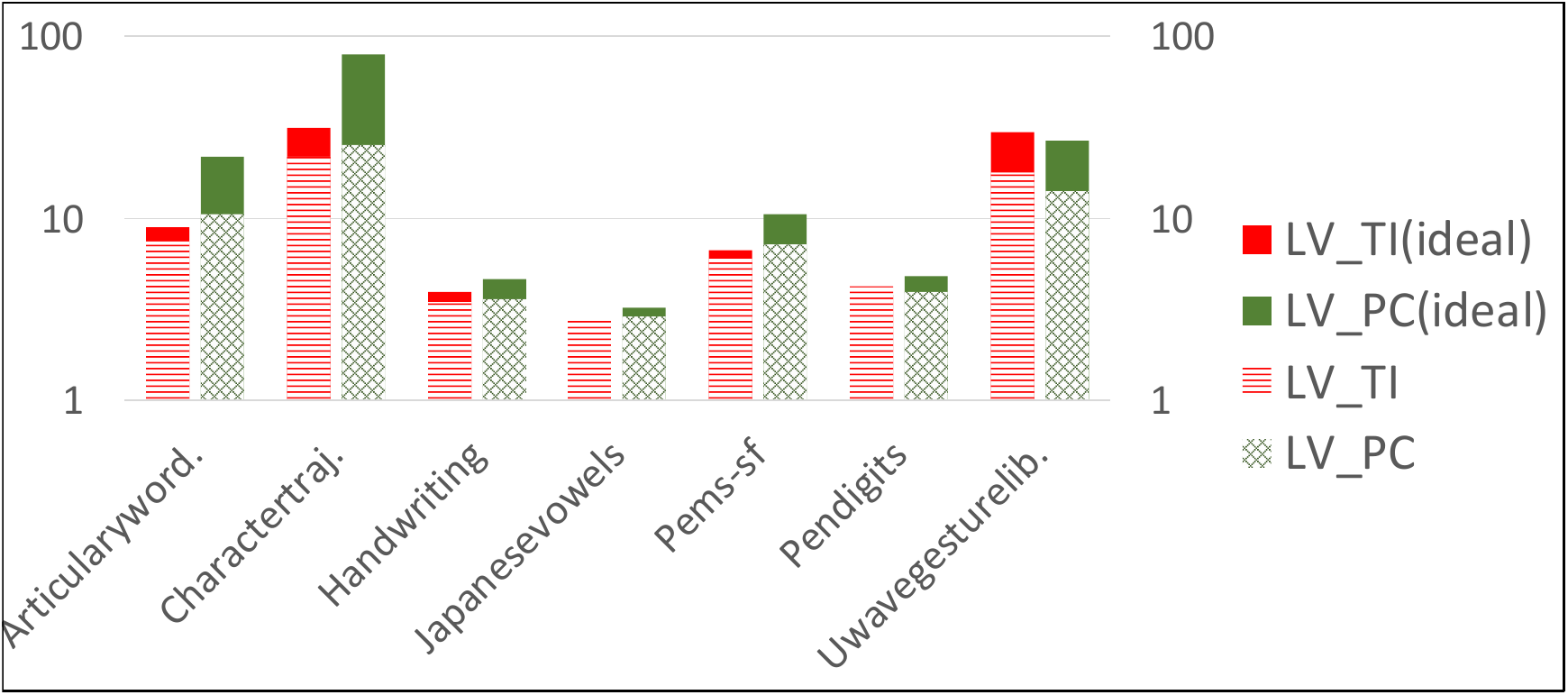}
    \caption{Speedups of LB\_TI and LB\_PC, and their ideal results when all runtime overhead is removed.}
    \label{fig:overhead}
\end{figure}

\mypar{Time overhead} To study the impact from the runtime overhead, Figure~\ref{fig:overhead} puts the speedups of LB\_TI and LB\_PC in the backdrop of their ideal speedups where runtime overhead is excluded. (For readability, we include only some of the datasets in that graph.) The vertical axis is in log scale to make all the bars visible. The largest gaps appear on datasets \texttt{Charactertrajectories} and \texttt{Uwavegesturelibrary}, with a 10-55$\times$ speedup left untapped due to the runtime overhead, indicating that further optimization of the implementations of the two methods could potentially yield more speedups. 

\mypar{Window size} The bottom half of Table~\ref{tab:online} shows the results when the window size is 10 (again, capped at the length of a time series). As window size reduces, all methods become more effective in finding skip opportunities with the average skips increasing to over 60\% in all the methods. The three proposed methods, LB\_TI, LB\_PC, and TC-DTW, still show a clear edge over LB\_MV.

\mypar{Dimensions} As mentioned earlier, even though some of the datasets have points with a high dimension, in practice, practitioners typically first use dimension reduction methods to lower the dimensions to a small number~\cite{Hu+:ICDM2013} to avoid the so-called curse of dimensionality. In all the literature on multivariate DTW applications that we have checked, the largest dimension used is less than 10. Therefore in our examination of the impact of dimensionality on the methods, we focus on three settings: $dim$=3,5,10. Our experiments run on the datasets that have dimensions greater than 8 by changing the used dimensions to the first $dim$ dimensions of the data. Table~\ref{tab:dims} reports the speedups from the methods over the default window-based DTW. All methods see some reductions of speedups as the number of dimensions increase. It is intuitive: As data become sparser in higher-dimension spaces, the bounding boxes become larger and lower bounds become looser.  Nevertheless, TC-DTW still consistently outperforms LB\_MV on all the datasets and dimensions. 

\mypar{Performance cross machines} To examine whether the new methods consistently show benefits across different machines, besides the experiments on the AMD machine, we have measured the performance of the methods on the Intel machine (configuration given in Section~\ref{sec:methodology}). Table~\ref{tab:intel} reports the results. The setting is identical to that in Table~\ref{tab:online} (W=40). Compared to the results on the AMD machine in Table~\ref{tab:online}, the speedups of the three methods are all somewhat smaller, reflecting the influence of the hardware differences in terms of the relative speeds between CPU and memory accesses. But the trends are the same. The results confirm that the better efficacy of the three new methods over LB\_MV holds across machines, with TC-DTW excelling at an average 6.29$\times$ speedups.  

Overall, the experiments confirm the advantages of the new methods over the current approach LB\_MV in accelerating multivariate DTW. The advantages consistently hold across datasets, data sizes, dimensions, window sizes, and machines. The advantages of the new methods over LB\_MV are especially pronounced on large datasets with large windows, where, time is especially of concern.

\section{Discussions}
\label{sec:discussions}

The two methods presented in the paper are the crystallized results of the numerous ideas we have explored. The other ideas are not as effective. We briefly describe one of them for its potential for future development. 

This idea is also about clustering, but instead of point-level clustering, it considers clustering of query series. The basic insight behind it is that the tight bounds from LB\_AD meet Triangle Inequality with the Euclidean distances between two query series. That is, $LB\_AD(Q_j, C) - ED(Q_i, Q_j) \le LB\_AD(Q_i, C)\le LB\_AD(Q_j, C) + ED(Q_i, Q_j)$, where $ED$ stands for Euclidean distance. We omit the proof here but note that it can be easily proved if one notices that such an inequality holds at every corresponding point between $Q_i$ and $Q_j$. Therefore, for two queries series similar to each other, we can use the inequality to estimate one's lower bounds with a candidate series based on those of the other series. We attempted with some quick ways to do clustering (e.g., Euclidean distances among sample points),  but did not observe significant benefits except on only a few time series. The reason is that even though the method can sometimes give bounds tighter than LB\_MV, the degree is too small to change the filtering result. Applying the idea to each segment of a series might give better results. We leave it for future exploration.

It is worth mentioning another possible application scenario of the proposed methods. In the scenario studied in our experiments, the bounding boxes and neighboring point distances are computed on the fly when a query series arrives. Another option is to switch the role of query and candidate series and compute the bounding boxes and neighboring distances of the candidate series ahead of time (e.g., when setting up the database). The pre-computed info can then be used in processing queries. Such an option incurs extra space cost as the bounding boxes and neighbor point distances need to be permanently stored but if that is not a concern, it could further amplify the performance benefits of the methods.

\section{Related Work}
\label{sec:related}

The most significant advantage of DTW is its invariance against signal
warping (shifting and scaling in the time axis, or Doppler effect). Experimental comparison of DTW to most other highly cited distance measures on many datasets concluded that DTW almost
always outperforms other measures~\cite{Wang+:DMKD2013}. Therefore, DTW has become one
of the most preferable measures in pattern matching tasks for time series data. It has been used in robotics, medicine~\cite{Chadwick+:DEST2011}, biometrics, speech processing~\cite{Adams+:ISMIR05}, climatology, aviation, gesture recognition~\cite{Alon+:PAMI2009}, user interfaces~\cite{Laerhoven+:ICMLA2009}, industrial processing, cryptanalysis~\cite{Dupasquier+:Chaos2011}, mining of historical manuscripts~\cite{Huber+:MVA2011}, space exploration,
wildlife monitoring, and so on~\cite{Inglada+:sense2015,Rak+:TKDD2013}.

DTW remains popular in the era of deep learning. DTW-based temporal data analytics are still widely used for its advantages in better interpretability over DNN. Moreover, there is strong interest in combining DTW with deep learning. The most common integration of DTW and deep learning is to use DTW as a loss function. Studies have shown that DTW, compared to conventional Euclidean distance loss, leads to a better performance in a number of applications~\cite{Cuturl+:arxiv2017} through softmin~\cite{Chang+:arxiv2019,Shah+:CODS16}.  A more recent way to combine DTW and deep learning is to use DTW as a replacement of kernels in convolutional neural networks, creating DTWNet that shows improved inference accuracy~\cite{Cai+:NIPS2019}. There are also some other integrations of DTW with DNN~\cite{Thiollire+:AHD2015}. In addition, there is interest in using DTW as a distance measure in learning time series shapelets~\cite{Shah+:CODS16}. Besides serving as a similarity measure, DTW could be leveraged as a feature extracting tool. For example, predefined patterns can be identified in the data via DTW
computing. Subsequently these patterns could be used to classify the temporal data into categories~\cite{Kate:DMKD16,Giraldo+:SIU18}.

Many studies have been conducted to enhance the speed of DTW-based analytics, but mostly for single-dimensional data, including the many lower bound functions for univariate time series, such as LB\_Yi~\cite{Yi+:VLDB2000}, LB\_Kim~\cite{Kim+:ICDE2001}, LB\_Keogh~\cite{Keogh+:VLDB2006}, LB\_Improved~\cite{Lemire+:PR2009} and LB\_ECorner~\cite{Zhou+:ICDE2011}, various methods to enable early abandoning~\cite{Rak+:TKDD2013}, the combination of hierarchy k-means and lower bounding~\cite{Tan+:SDM17}, and so on. 

There are hundreds of research efforts that use DTW in a multi-dimensional setting~\cite{Ridgely+:2009, Kela+:2006, Ko+:2005, Liu+:Mob2009, Peti+:TGRS12, Wang+:interspeech2013}. Research is however scarce in designing optimizations specifically for accelerating multivariate DTW, despite the even higher cost of multivariate DTW than univariate DTW. Almost all of the existing applications of multivariate DTW are based on a straightforward extension of the single-dimensional lower bound (LB\_Keogh [27]) to the multi-dimensional data~\cite{Rath+:2003}, described as LB\_MV in Section~\ref{sec:background}. The previously proposed optimizations have concentrated on the reduction or transformations of the high dimension space. Li and others~\cite{Li+:physica2015,Li+:elsevier2017} tries to reduce all dimensions to a single dimension (center series) and then use univariate methods; Hu and others have studied the effects of giving more emphasis on important dimensions~\cite{Hu+:ICDM2013}; Gong and others demonstrate that by rotating the space, one could improve the tightness of the lower bounds~\cite{Gong+:Springer2015}. Some optimizations proposed for univariate DTW could potentially benefit applications of multivariate DTW. For instance, in DTW-based Nearest Neighbor, the early abandoning strategy stops the computation of the DTW between a query sequence and a reference sequence if evidence shows that the DTW distance will definitely exceed the best-so-far. All of these prior studies have concentrated on factors outside the calculations of lower bounds for more effective filtering. This work gives the first systematic exploration on the optimizations of lower bounds calculations specific to multivariate DTW.  Being complementary to the previously proposed optimizations, it can be used with some of the existing optimizations together.

Triangle inequality has been leveraged in other data mining and machine learning optimizations (e.g., Yinyang K-Means~\cite{Ding+:ICML15}). We are not aware of a previous introduction of it in accelerating multivariate DTW. 

\section{Conclusions}
\label{sec:conclusions}

This paper has introduced Triangle Inequality and Pointer Clustering into the algorithm design of multivariate DTW. The incorporation of Triangle Inequality allows the use of scalar computations to replace vector distance calculations in computing tighter lower bounds; the integration of Point Clustering allows the use of quantization-based clustering to tighten the lower bounds. It also presents several design optimizations to the two methods, including periodical distance calculations, adaptive quantization, uniform window grouping, and data-adaptive selection. Experiments on 13 datasets show that the resulting method, TC-DTW, consistently outperforms the current most commonly used method LB\_MV significantly, across datasets, data sizes, dimensions, window sizes, and hardware, suggesting the potential of TC-DTW as a drop-in replacement for LB\_MV, a method that has prevailed in the most part of the last two decades. 

\balance
\bibliographystyle{plain}

\begin{thebibliography}{10}

\bibitem{Adams+:ISMIR05}
N.~Adams, D.~Marquez, and G.~Wakefield.
\newblock Iterative deepening for melody alignment and retrieval.
\newblock In {\em Proceedings of ISMIR}, pages 199--206, 2005.

\bibitem{Alon+:PAMI2009}
Jonathan Alon, Vassilis Athitsos, Quan Yuan, and Stan Sclaroff.
\newblock A unified framework for gesture recognition and spatiotemporal
  gesture segmentation.
\newblock {\em {IEEE} Trans. Pattern Anal. Mach. Intell.}, 31(9):1685--1699,
  2009.

\bibitem{Bagnall+:arxiv2018}
Anthony Bagnall, Hoang~Anh Dau, Jason Lines, Michael Flynn, James Large, Aaron
  Bostrom, Paul Southam, and Eamonn Keogh.
\newblock The uea multivariate time series classification archive, 2018.
\newblock In {\em Arxiv}, 2018.

\bibitem{Cai+:NIPS2019}
Xingyu Cai, Tingyang Xu, Jinfeng Yi, Junzhou Huang, and Sanguthevar
  Rajasekaran.
\newblock Dtwnet: a dynamic time warping network.
\newblock In H.~Wallach, H.~Larochelle, A.~Beygelzimer, F.~d\textquotesingle
  Alch\'{e}-Buc, E.~Fox, and R.~Garnett, editors, {\em Advances in Neural
  Information Processing Systems 32}, pages 11640--11650. Curran Associates,
  Inc., 2019.

\bibitem{Chadwick+:DEST2011}
N.~A. Chadwick, D.~A. McMeekin, and T.~Tan.
\newblock Classifying eye and head movement artifacts in eeg signals.
\newblock In {\em Proceedings of IEEE DEST}, pages 285--291, 2011.

\bibitem{Chang+:arxiv2019}
Chien{-}Yi Chang, De{-}An Huang, Yanan Sui, Li~Fei{-}Fei, and Juan~Carlos
  Niebles.
\newblock {D3TW:} discriminative differentiable dynamic time warping for weakly
  supervised action alignment and segmentation.
\newblock In {\em {IEEE} Conference on Computer Vision and Pattern Recognition,
  {CVPR} 2019, Long Beach, CA, USA, June 16-20, 2019}, pages 3546--3555.
  Computer Vision Foundation / {IEEE}, 2019.

\bibitem{Cuturl+:arxiv2017}
Marco Cuturi and Mathieu Blondel.
\newblock Soft-dtw: a differentiable loss function for time-series.
\newblock In Doina Precup and Yee~Whye Teh, editors, {\em Proceedings of the
  34th International Conference on Machine Learning, {ICML} 2017, Sydney, NSW,
  Australia, 6-11 August 2017}, volume~70 of {\em Proceedings of Machine
  Learning Research}, pages 894--903. {PMLR}, 2017.

\bibitem{UCRArchive2018}
Hoang~Anh Dau, Eamonn Keogh, and Kaveh~Kamgar et~al.
\newblock The ucr time series classification archive, October 2018.
\newblock \url{https://www.cs.ucr.edu/~eamonn/time_series_data_2018/}.

\bibitem{Ding+:ICML15}
Y.~Ding, Y.~Zhao, X.~Shen, M.~Musuvathi, and T.~Mytkowicz.
\newblock Yinyang k-means: A drop-in replacement of the classic k-means with
  consistent speedup.
\newblock In {\em Proceedings of the 32nd International Conference on Machine
  Learning}, 2015.

\bibitem{Dupasquier+:Chaos2011}
B.~Dupasquier and S.~Burschka.
\newblock Data mining for hackers–encrypted traffic mining.
\newblock In {\em Proceedings of the 28th Chaos Comm' Congress}, 2011.

\bibitem{Kumar2006}
A~Kumar et~al.
\newblock Duration of hypotension before initiation of effective antimicrobial
  therapy is the critical determinant of survival in human septic shock.
\newblock {\em Critical care medicine}, 2006.
\newblock DOI: 10.1097/01.CCM.0000217961.75225.E9.

\bibitem{Giraldo+:SIU18}
Sergio~I. Giraldo, Ariadna Ortega, Alfonso P{\'{e}}rez, Rafael Ram{\'{\i}}rez,
  George Waddell, and Aaron Williamon.
\newblock Automatic assessment of violin performance using dynamic time warping
  classification.
\newblock In {\em 26th Signal Processing and Communications Applications
  Conference, {SIU} 2018, Izmir, Turkey, May 2-5, 2018}, pages 1--3. {IEEE},
  2018.

\bibitem{Gong+:Springer2015}
Xudong Gong, Yan Xiong, Wenchao Huang, Lei Chen, Qiwei Lu, and Yiqing Hu.
\newblock Fast similarity search of multi-dimensional time series via segment
  rotation.
\newblock In Matthias Renz, Cyrus Shahabi, Xiaofang Zhou, and Muhammad~Aamir
  Cheema, editors, {\em Database Systems for Advanced Applications - 20th
  International Conference, {DASFAA} 2015, Hanoi, Vietnam, April 20-23, 2015,
  Proceedings, Part {I}}, volume 9049 of {\em Lecture Notes in Computer
  Science}, pages 108--124. Springer, 2015.

\bibitem{Hu+:ICDM2013}
Bing Hu, Yanping Chen, Jesin Zakaria, Liudmila Ulanova, and Eamonn~J. Keogh.
\newblock Classification of multi-dimensional streaming time series by
  weighting each classifier's track record.
\newblock In Hui Xiong, George Karypis, Bhavani~M. Thuraisingham, Diane~J.
  Cook, and Xindong Wu, editors, {\em 2013 {IEEE} 13th International Conference
  on Data Mining, Dallas, TX, USA, December 7-10, 2013}, pages 281--290. {IEEE}
  Computer Society, 2013.

\bibitem{Huber+:MVA2011}
Reinhold Huber{-}M{\"{o}}rk, Sebastian Zambanini, Maia Zaharieva, and Martin
  Kampel.
\newblock Identification of ancient coins based on fusion of shape and local
  features.
\newblock {\em Mach. Vis. Appl.}, 22(6):983--994, 2011.

\bibitem{Inglada+:sense2015}
Jordi Inglada, Marcela Arias, Benjamin Tardy, Olivier Hagolle, Silvia Valero,
  David Morin, G{\'{e}}rard Dedieu, Guadalupe Sepulcre, Sophie Bontemps, Pierre
  Defourny, and Benjamin Koetz.
\newblock Assessment of an operational system for crop type map production
  using high temporal and spatial resolution satellite optical imagery.
\newblock {\em Remote. Sens.}, 7(9):12356--12379, 2015.

\bibitem{Kate:DMKD16}
Rohit~J. Kate.
\newblock Using dynamic time warping distances as features for improved time
  series classification.
\newblock {\em Data Min. Knowl. Discov.}, 30(2):283--312, 2016.

\bibitem{Kela+:2006}
Juha Kela, Panu Korpip{\"{a}}{\"{a}}, Jani M{\"{a}}ntyj{\"{a}}rvi, Sanna
  Kallio, Giuseppe Savino, Luca Jozzo, and Sergio~Di Marca.
\newblock Accelerometer-based gesture control for a design environment.
\newblock {\em Personal and Ubiquitous Computing}, 10(5):285--299, 2006.

\bibitem{Keogh+:VLDB2006}
Eamonn~J. Keogh, Li~Wei, Xiaopeng Xi, Sang{-}Hee Lee, and Michail Vlachos.
\newblock Lb{\_}keogh supports exact indexing of shapes under rotation
  invariance with arbitrary representations and distance measures.
\newblock In Umeshwar Dayal, Kyu{-}Young Whang, David~B. Lomet, Gustavo Alonso,
  Guy~M. Lohman, Martin~L. Kersten, Sang~Kyun Cha, and Young{-}Kuk Kim,
  editors, {\em Proceedings of the 32nd International Conference on Very Large
  Data Bases, Seoul, Korea, September 12-15, 2006}, pages 882--893. {ACM},
  2006.

\bibitem{Kim+:ICDE2001}
Sang{-}Wook Kim, Sanghyun Park, and Wesley~W. Chu.
\newblock An index-based approach for similarity search supporting time warping
  in large sequence databases.
\newblock In Dimitrios Georgakopoulos and Alexander Buchmann, editors, {\em
  Proceedings of the 17th International Conference on Data Engineering, April
  2-6, 2001, Heidelberg, Germany}, pages 607--614. {IEEE} Computer Society,
  2001.

\bibitem{Laerhoven+:ICMLA2009}
Kristof~Van Laerhoven, Eugen Berlin, and Bernt Schiele.
\newblock Enabling efficient time series analysis for wearable activity data.
\newblock In M.~Arif Wani, Mehmed~M. Kantardzic, Vasile Palade, Lukasz~A.
  Kurgan, and Yuan~(Alan) Qi, editors, {\em International Conference on Machine
  Learning and Applications, {ICMLA} 2009, Miami Beach, Florida, USA, December
  13-15, 2009}, pages 392--397. {IEEE} Computer Society, 2009.

\bibitem{Lemire+:PR2009}
Daniel Lemire.
\newblock Faster retrieval with a two-pass dynamic-time-warping lower bound.
\newblock {\em Pattern Recognit.}, 42(9):2169--2180, 2009.

\bibitem{Li+:physica2015}
Hailin Li.
\newblock Piecewise aggregate representations and lower-bound distance
  functions for multivariate time series.
\newblock {\em Physica}, 427(A):10--25, 2015.

\bibitem{Li+:elsevier2017}
Hailin Li.
\newblock Distance measure with improved lower bound for multivariate time
  series.
\newblock {\em Elsevier}, 468(C):622--637, 2017.

\bibitem{Liu+:Mob2009}
Jiayang Liu, Lin Zhong, Jehan Wickramasuriya, and Venu Vasudevan.
\newblock uwave: Accelerometer-based personalized gesture recognition and its
  applications.
\newblock {\em Pervasive Mob. Comput.}, 5(6):657--675, 2009.

\bibitem{Ko+:2005}
Ko~MH, West G, Venkatesh S, and Kumar M.
\newblock Online context recognition in multisensor systems using dynamic time
  warping.
\newblock In {\em Proceedings of the IEEE international conference on
  intelligent sensors, sensor networks and information processing (ISSNIP)},
  page 283–288, 2005.

\bibitem{Peti+:TGRS12}
Fran{\c{c}}ois Petitjean and Jonathan Weber.
\newblock Efficient satellite image time series analysis under time warping.
\newblock {\em {IEEE} Geosci. Remote. Sens. Lett.}, 11(6):1143--1147, 2014.

\bibitem{Rak+:TKDD2013}
Thanawin Rakthanmanon, Bilson J.~L. Campana, Abdullah Mueen, Gustavo E. A.
  P.~A. Batista, M.~Brandon Westover, Qiang Zhu, Jesin Zakaria, and Eamonn~J.
  Keogh.
\newblock Addressing big data time series: Mining trillions of time series
  subsequences under dynamic time warping.
\newblock {\em {ACM} Trans. Knowl. Discov. Data}, 7(3):10:1--10:31, 2013.

\bibitem{Rath+:2003}
Toni Rath and R.~Manmatha.
\newblock Lower-bounding of dynamic time warping distances for multivariate
  time series.
\newblock 02 2003.

\bibitem{Ridgely+:2009}
Ridgely~Robert S and Tudor~G Field.
\newblock Guide to the songbirds of south america the passerines.
\newblock {\em Mildred Wyatt-Wold Series in Ornithology}, 2009.

\bibitem{SakoeBand}
H.~Sakoe and S.~Chiba.
\newblock Dynamic programming algorithm optimization for spoken word.
\newblock {\em Trans. Acoustics, Speech, and Signal Proc.}, ASSP-26:43--49,
  1978.

\bibitem{Shah+:CODS16}
Mit Shah, Josif Grabocka, Nicolas Schilling, Martin Wistuba, and Lars
  Schmidt{-}Thieme.
\newblock Learning dtw-shapelets for time-series classification.
\newblock In Madhav Marathe, Mukesh~K. Mohania, Mausam, and Prateek Jain,
  editors, {\em Proceedings of the 3rd {IKDD} Conference on Data Science,
  {CODS} 2016, Pune, India, March 13-16, 2016}, pages 3:1--3:8. {ACM}, 2016.

\bibitem{Shokoohi+:DMKD17}
Mohammad Shokoohi{-}Yekta, Bing Hu, Hongxia Jin, Jun Wang, and Eamonn~J. Keogh.
\newblock Generalizing {DTW} to the multi-dimensional case requires an adaptive
  approach.
\newblock {\em Data Min. Knowl. Discov.}, 31(1):1--31, 2017.

\bibitem{Tan+:SDM17}
{Chang Wei} Tan, {Geoffrey I.} Webb, and Francois Petitjean.
\newblock Indexing and classifying gigabytes of time series under time warping.
\newblock In {Nitesh } Chawla and {Wei } Wang, editors, {\em Proceedings of the
  17th SIAM International Conference on Data Mining}, pages 282--290. Society
  for Industrial \& Applied Mathematics (SIAM), January 2017.
\newblock SIAM International Conference on Data Mining 2017, SDM 2017 ;
  Conference date: 27-04-2017 Through 29-04-2017.

\bibitem{Thiollire+:AHD2015}
Roland Thiolli{\`e}re, Ewan Dunbar, Gabriel Synnaeve, Maarten Versteegh, and
  Emmanuel Dupoux.
\newblock A hybrid dynamic time warping-deep neural network architecture for
  unsupervised acoustic modeling.
\newblock In {\em INTERSPEECH}, 2015.

\bibitem{Wang+:interspeech2013}
Jun Wang, Arvind Balasubramanian, Luis~Mojica de~La~Vega, Jordan~R. Green,
  Ashok Samal, and Balakrishnan Prabhakaran.
\newblock Word recognition from continuous articulatory movement time-series
  data using symbolic representations.
\newblock In Jan Alexandersson, Peter Ljungl{\"{o}}f, Kathleen~F. McCoy,
  Fran{\c{c}}ois Portet, Brian Roark, Frank Rudzicz, and Michel Vacher,
  editors, {\em Proceedings of the Fourth Workshop on Speech and Language
  Processing for Assistive Technologies, {SLPAT} 2013, Grenoble, France, August
  21-22, 2013}, pages 119--127. Association for Computational Linguistics,
  2013.

\bibitem{Wang+:DMKD2013}
Xiaoyue Wang, Abdullah Mueen, Hui Ding, Goce Trajcevski, Peter Scheuermann, and
  Eamonn~J. Keogh.
\newblock Experimental comparison of representation methods and distance
  measures for time series data.
\newblock {\em Data Min. Knowl. Discov.}, 26(2):275--309, 2013.

\bibitem{Yi+:VLDB2000}
Byoung{-}Kee Yi and Christos Faloutsos.
\newblock Fast time sequence indexing for arbitrary lp norms.
\newblock In Amr~El Abbadi, Michael~L. Brodie, Sharma Chakravarthy, Umeshwar
  Dayal, Nabil Kamel, Gunter Schlageter, and Kyu{-}Young Whang, editors, {\em
  {VLDB} 2000, Proceedings of 26th International Conference on Very Large Data
  Bases, September 10-14, 2000, Cairo, Egypt}, pages 385--394. Morgan Kaufmann,
  2000.

\bibitem{Zhou+:ICDE2011}
Mi~Zhou and Man~Hon Wong.
\newblock Boundary-based lower-bound functions for dynamic time warping and
  their indexing.
\newblock {\em Inf. Sci.}, 181(19):4175--4196, 2011.

\end{thebibliography}

\end{document}